\documentclass{article}
\usepackage{iclr2027_conference}
\newif\ifarxiv\arxivtrue   
\ifarxiv\iclrfinalcopy\fi

\usepackage{amsmath,amsfonts,bm}

\def\eqref#1{equation~\ref{#1}}

\def\1{\bm{1}}

\DeclareMathAlphabet{\mathsfit}{\encodingdefault}{\sfdefault}{m}{sl}
\SetMathAlphabet{\mathsfit}{bold}{\encodingdefault}{\sfdefault}{bx}{n}

\usepackage{natbib}
\usepackage{hyperref}
\usepackage{url}
\usepackage{booktabs}
\usepackage{amsmath,amssymb,amsfonts,amsthm}
\usepackage{graphicx}
\usepackage{subcaption}
\usepackage{microtype}
\usepackage{xcolor}
\usepackage{colortbl}
\usepackage{algorithm}
\usepackage{algorithmic}
\usepackage{booktabs}       
\usepackage{threeparttable} 
\usepackage{placeins}
\usepackage{xspace}

\newtheorem{theorem}{Theorem}[section]

\newtheorem{proposition}[theorem]{Proposition}

\theoremstyle{definition}

\newif\ifmarkchanges\markchangestrue
\ifmarkchanges\else\fi

\title{Multi-agent Scaling  Across Disjunctive and   Compensatory Tasks}

\author{Carolina Fortuna, Blaž Bertalanič}

\begin{document}
\maketitle
\ifarxiv\lhead{Preprint}\fi

\begin{abstract}

Multi-agent LLM systems are often expected to improve as team size increases, yet the scaling behavior may depend on task structure. \textit{Our central contribution is to introduce Steiner's taxonomy of group tasks as a framework for analyzing multi-agent LLM scaling} and focusing the analysis on  disjunctive and compensatory tasks. We model independently sampled agents as conditionally independent given the item, which yields their large-team limits: plurality voting converges to the model's modal answer, and averaging converges to the model's item-level bias. Across selected representative benchmarks, 13 open-weight models, and teams of up to 30 agents, we find qualitatively different scaling behavior. On disjunctive tasks, the probability that at least one agent is correct grows by 5–20 points with team size, but plurality voting over agents
that answer directly realises almost none of this potential, as the model predicts to within 0.5 points on average. Multi-round revision raises accuracy considerably, yet the gain is nearly the same with one peer as with 29. In contrast, scaling provides little benefit on Fermi estimation, despite its natural suitability for aggregation: item-level biases shared across the samples of a model account for about 87\% of the squared error, so averaging reduces error by only about 6\%. Combining model families helps on Fermi estimation but does not  surpass  the strongest member on disjunctive tasks. These results show that task structure, together with the mechanism combining member outputs, is a  fundamental determinant of team scaling. [Open code placeholder]

\end{abstract}

\section{Introduction}
\textbf{Motivation.} With the rapid emergence of Large Language Model (LLM) multi-agent frameworks \citep{wu2023autogen, li2023camel, du2023improving}, researchers have deployed LLM ensembles and multi-agent debates across a multitude of reasoning and decision-making problems under the implicit assumption that scaling team size $N$ universally enhances system performance. As LLMs are trained on human generated knowledge largely encoded through language, that inherently also captures cultural and behavioral traits, we turn to observing multi-agent behavior in a similar way human group behavior was observed and studied.  The ability of a group to outperform its individual members is among the most studied phenomenon in cognitive science, economics, and social psychology. In his classic 1907  experiment, Francis Galton observed that while individual fairgoers made highly scattered  
guesses regarding the weight of an ox, the median of their estimates fell within $0.8\%$ of the true weight \citep{galton1907wisdom, surowiecki2004wisdom}. Conversely, the foundational Ringelmann effect \citep{ingham1974ringelmann} reveals the opposing tension in collective    dynamics: as group size increases, teams frequently suffer from diminishing per-member returns and performance saturation. 
    
In Ivan Steiner's seminal taxonomy of group processes \citep{steiner1972group}, resolving whether a collective achieves Galtonian crowd wisdom or succumbs to Ringelmann-style process loss depends  on the underlying task structure. The tasks observed by Galton are the so-called \textbf{compensatory tasks}  \cite{steiner1972group}  where group performance is determined by statistical aggregation of continuous estimates.  In human crowds, because individual participants possess idiosyncratic cognitive biases and independent heuristic noise, their random errors  compensate for one another \citep{larrick2006intuitions,davisstober2014crowd}. 

Ringelmann originally observed this performance drop on \textbf{additive tasks}, where per-member productivity decayed as group size increased due to coordination losses and social loafing \citep{latane1979many, ingham1974ringelmann}. It was later shown by \citet{kerr1983dispensability} that group performance losses are fundamentally mediated by Steiner's task typology and the perceived dispensability of individual effort. While additive and \textbf{conjunctive} tasks frequently trigger free-rider effects or coordination bottlenecks, \textbf{compensatory tasks}  can circumvent these losses when members provide independent, unbiased estimates that aggregate symmetrically. This established that collective human team scaling laws are fundamentally \textbf{task-dependent}. The same observation has been recently made by ~\cite{kim2025towards}: LLM agent team performance is task dependent. At the same time, inspired by Ringlemann's findings, \cite{bertalanivc2026ringelmann} developed a scaling law for effective team size.

\textbf{Prior art.} Without classifying them into the Steiner taxonomy nor recognizing them explicitly as disjunctive or compensatory tasks, the community has already investigated related multi-agent performance. The rich state of the art on disjunctive tasks represented by benchmarks such as MMLU, GSM8K, MATH500, ARC reports  gains from sampling and voting \citep{wang2022self, li2024moreagents}, and from debate \citep{du2023improving} although these gains can saturate or reverse as the number of calls grows \citep{chen2024morecalls,bertalanivc2026ringelmann}. 
Debate induces conformity \citep{weng2025conformity}, and correlated errors limit ensembles \citep{kim2025correlated}. 
 Improvements have also been reported for compensatory tasks represented by Fermi \cite{fermi2021realfp, liu2025fermi} and guesstimate \cite{chuang2025probing} problems.  The closest related works are \citet{chen2024morecalls}, who explain non-monotone voting by a mixture of easy and hard queries, the basis of our plurality limit, and \citet{bertalanivc2026ringelmann}, who fit a law for the effective team size $N_{\text{eff}}/N$ across 44 conditions and find that only heterogeneous teams escape hard ceilings; homogeneous teams also saturate in \citet{yang2026diversity}, and LLM teams fall short of their best member in \citet{pappu2026experts}.

\textbf{This work.} \textit{Our central contribution is to introduce Steiner's taxonomy of group tasks as a framework for analyzing multi-agent LLM scaling} enabling the  formalization and characterization of multi-agent performance on formal  task types:

\textbf{1.} We connect benchmarks extensively used the evaluation of LLM systems to the Steiner taxonomy and mathematically formalise agent behaviour on these tasks, including the large-team limits of independently sampled agents (Section~\ref{sec:theory}).

\textbf{2.} We investigate multi-agent scaling with 13 open-weight models and teams of up to 30 agents, analysing the independence and diversity assumptions (Section~\ref{sec:scaling_dynamics}). On disjunctive tasks, pass@$N$ grows steadily with team size, whereas plurality voting converges to the model's modal answer, as predicted to within 0.5 points by a conditional-independence model. Revision rounds raise accuracy, but almost equally with one peer or 29 (Section  \ref{sec:disjunctive_ceiling}).

\textbf{3.} We show that unbiasedness fails for compensatory Fermi estimate  tasks at the level of individual items: item-level bias accounts for about 87\% of the squared log-error, so averaging reduces error by only about 6\%; part of the apparent cross-model correlation is due to implausible reference labels (Section~\ref{sec:compensatory_ceiling}).

\textbf{4.} We show that heterogeneous teams improve substantially on their average member but, on disjunctive tasks, don't exceed their strongest member; on the compensatory Fermi estimation, a team of five 7B--8B models outperforms its members (Sections~\ref{sec:disjunctive_ceiling} and~\ref{sec:compensatory_ceiling}).

\section{Taxonomy and Theoretical Scaling Framework}
\label{sec:theory}

\subsection{Steiner's Task Taxonomy and Modern LLM Benchmarks}
\label{sec:steiner_taxonomy}

We map representative benchmarks  used to evaluate LLM-based multi-agent teams onto Steiner's task types \citep{steiner1972group}, according to the rule by which a multi-agent system combines member contributions (Table~\ref{tab:steiner_taxonomy}); the mapping describes a combination rule rather than an intrinsic property of a benchmark. The  remainder of the paper focuses on disjunctive tasks, which depend on the skill of individual members, and compensatory tasks, which depend on error cancellation; we leave the other task types for future work.

\textbf{Disjunctive Tasks} in which  the outcome is established through $Y = \max_i \{y_i\}$ or plurality/majority/verifier selection and/or debate. The tasks are unitary and optimizing and group success requires that \emph{at least one} agent discovers a valid solution, and the group then recognizes it as such. Examples include  mathematical reasoning, multiple-choice knowledge and science questions, and code generation (Table~\ref{tab:steiner_taxonomy}), which MASes address through plurality voting \citep{wang2022self}, verifier re-ranking, and debate \citep{du2023improving, wang2025mixture}.

\begin{table}[t]
\caption{Categorization of  tasks according to the Steiner taxonomy. Disjunctive and compensatory tasks are briefly discussed in this section; with additional details on all tasks in Appendix~\ref{sec:other_cells}.}
\label{tab:steiner_taxonomy}
\begin{center}
\footnotesize
\resizebox{\textwidth}{!}{
\begin{tabular}{p{1.0cm}p{1.8cm}p{1.5cm}p{6.0cm}p{3.6cm}}
\toprule
\textbf{Task} & \textbf{Combination Operator} & \textbf{Output Domain} & \textbf{Top-Conference Benchmarks (2024--2026)} & \textbf{LLM Aggregation Mechanism}  \\
\midrule
\textbf{Dis- junc- tive} & $Y = \max_i \{y_i\}$ \newline (or plurality) & Discrete ($\mathcal{Y}$) & MATH-500, GSM8K, MMLU-Pro \citep{wang2024mmlu}, GPQA \citep{rein2024gpqa}, BigCodeBench \citep{zhuo2024bigcodebench}, ARC & Plurality voting, verifier selection, multi-agent debate \\
\addlinespace
\textbf{Com- pen- satory} & $Y = \left(\prod_{i=1}^n x_i\right)^{\frac{1}{n}}$ \newline (or median) & Continuous ($\mathbb{R}^+, [0,1]$) & ForecastBench \citep{forecastbench2024}, Guesstimate \citep{chuang2025probing}, RealFP \citep{fermi2021realfp}, SynthFP, SciOly & Geometric/arithmetic mean, median, calibrated Brier aggregation  \\
\addlinespace
\textbf{Addi- tive} & $Y = \sum_i y_i$ \newline or $\bigcup_i S_i$ & Cumulative sets & DAT \citep{olson2021naming}, open-ended ideation, parallel instruction synthesis & Deduplicated set union, parallel greedy generation  \\
\addlinespace
\textbf{Con- junc- tive} & $Y = \min_i \{y_i\}$ & Boolean / state chain & SWE-bench \citep{jimenez2024swebench}, TheAgentCompany \citep{theagentcompany2025}, OSWorld \citep{xie2024osworld}, BrowseComp \citep{kim2025towards} & When modeled as strict sequential pipeline, multi-step dependency verification  \\
\addlinespace
\textbf{Dis- cre- tionary} & $Y = f_{\text{group}}(\{y_i\})$ Dynamic & Heteroge- neous & AgentBench \citep{liu2023agentbench}, ChatEval \citep{chan2023chateval}, MoA \citep{wang2025mixture} & Meta-agent router, judge arbitration, dynamic role assignment \\
\bottomrule
\end{tabular}
}
\end{center}
\end{table}

\textbf{Compensatory Tasks} in which the outcome is determined by $Y = \text{Aggr}(x_1, \dots, x_n)$ via  statistical aggregation (geometric mean, arithmetic mean, or median) of continuous numerical estimates ($x_i \in \mathbb{R}^+$) or calibrated probabilities ($p_i \in [0, 1]$).  xamples include forecasting \citep{forecastbench2024}, guesstimation \citep{chuang2025probing}, and Fermi estimation \citep{fermi2021realfp, liu2025fermi}.

\subsection{Modeling Disjunctive Tasks Through Voting}
\label{sec:disjunctive_model}

\subsubsection{The Pure Disjunctive Case}
In Steiner's classical disjunctive framework \citep{steiner1972group}, a task is disjunctive if group success requires only that \emph{at least one} team member generates the correct solution $y^* \in \mathcal{Y}$ and that the
group recognises it. Let $p \in (0, 1)$ denote the probability that an individual agent independently produces the correct answer. For independent agents and an ideal verifier, the probability of collective success is $P_{\text{disj}}(N) = 1 - (1 - p)^N$, so the collective failure probability decays exponentially with team size $N$:  $P(\text{failure}) = (1 - p)^N \to 0$ when $N  \to \infty$. We refer to this, estimated empirically, as the proportion of items on which at least one agent is correct, as pass@$N$.

\subsubsection{Majority Voting and the Condorcet Jury Theorem}
In the absence of an oracle verifier, multi-agent frameworks aggregate discrete candidate outputs through majority or plurality voting \citep{wang2022self, du2023improving}. For a binary decision between alternative solutions, let each agent cast an independent vote $V_i \sim \text{Bernoulli}(p)$ for the correct solution $y^*$, where $p > \frac{1}{2}$. The majority vote decision $\hat{y}_{\text{maj}} =  \sum_{i=1}^n V_i > \frac{n}{2}$ succeeds with probability $\sum_{k>N/2}\binom{N}{k} p^k (1 - p)^{N - k}$. By Hoeffding's inequality under Bernoulli random variables where $a_i=0, b_i=1$ and deviation $t=p-\frac{1}{2}$, the probability that the majority vote errors is strictly upper-bounded by:
\begin{equation}
\label{eq:disjunctive_hoeffding}
    P(\hat{y}_{\text{maj}} \neq  y^*) \le \exp\left( -2n \left( p - 1/2 \right)^2 \right)
\end{equation}
Under uncorrelated agent assumption, Eq.~\ref{eq:disjunctive_hoeffding} shows that for agent accuracies  strictly greater than random guessing ($p > 0.5$), voting ensembles are bounded by  exponential error suppression as team size $N$ increases.

\subsection{Modeling Compensatory Tasks through Geometric Aggregation}
\label{sec:basic_model}

Next, we contrast the disjunctive voting paradigm with the classic statistical model of group judgment on \textbf{compensatory tasks} \citep{galton1907wisdom, surowiecki2004wisdom, steiner1972group}. Because continuous estimation tasks (e.g., Fermi problems) span many orders of magnitude, errors are evaluated in logarithmic space \citep{fermi2021realfp}. Let $g \in \mathbb{R}^+$ be the positive ground truth, and $x_i \in \mathbb{R}^+$ be agent $i$'s estimate for $i \in \{1, \dots, N\}$. The signed log-error is $e_i = \log_{10}\left( x_i/g \right)$. The team estimate $\bar{X}_N$ is formed via geometric mean aggregation, corresponding to arithmetic mean in log-error space:
\begin{equation}
    \label{eq:compensatory_log_mean}
    \bar{e}_N = \log_{10}\left( \frac{\bar{X}_N}{g} \right) = \frac{1}{N} \sum_{i=1}^N e_i
\end{equation}
\subsubsection{The Statistical Principle of Team Superiority}
To show why an averaged team judgment is expected to be more accurate than a typical individual \citep{gigone1997proper, armstrong2001principles}, we analyse the expected value and the variance of the collective error $\bar{e}_N$. In modelling human team performance, the theory relies on i) unbiased estimates where on average, individual human errors balance out to zero across the crowd: $\mathbb{E}[e_i] = 0 \quad \text{for all } i \in \{1, \dots, N\}$ and ii) independence where team members do not influence one another, meaning individual errors are statistically uncorrelated: $     \text{Cov}(e_i, e_j) = 0 \quad \text{for all } i \neq j$. If each individual's judgment error has a variance of $\text{Var}(e_i) = \sigma^2$, the variance of the team's collective estimate is calculated as:$\text{Var}(\bar{e}_N) = \frac{1}{N^2} \sum_{i=1}^N \text{Var}(e_i) = \sigma^2/N$, so that the standard error of the team's judgment shrinks at the rate $\text{SE}(\bar{e}_N) = \sigma/\sqrt{N}$. As team size $N$ grows, the variance approaches zero ($\lim_{N \to \infty} \text{Var}(\bar{e}_N) = 0$). This shows that the averaged team judgment will systematically be more accurate and consistent than any single individual judgment \citep{clemen1989combining, page2008difference}.
\subsection{Effective Team Capacity and Scaling Limits Across Task Types}
\label{sec:effective_capacity}

Comparing the mathematical foundations of the ideal disjunctive model (Section~\ref{sec:disjunctive_model}) and the ideal compensatory model (Section~\ref{sec:basic_model}) reveals the theoretical origin of the scaling dichotomy: the failure probability of independent agents decreases exponentially in disjunctive tasks while it drops polynomially in compensatory tasks. More details of what behavior is expected from correlated and biased agents on  disjunctive and compensatory task types are available in Appendix \ref{sec:ideal_dichotomy}. For correlated draws (i.e. answers sampled from a language model), the Kish design effect \citep{kish1965survey} provides the effective team capacity $N_{\text{eff}}$ defined as the size of an ideal, independent ensemble ($\rho = 0$) that achieves the equivalent variance reduction as a correlated ensemble of size $N$: $ N_{\text{eff}}(N) = N/(1 + (N-1)\rho)$. For any positive pairwise error correlation $\rho > 0$: $
    \lim_{n \to \infty} N_{\text{eff}}(N) = 1/\rho$.

An important consequence arises when agents are independent samples from the same model, as in our experiments. Conditional on the item $k$, the outputs of such agents are independent and identically distributed with an item-specific distribution $P_k$. The pairwise correlation of their correctness across items then equals $\rho = \mathrm{Var}_k(p_k)/(\bar p(1-\bar p))$, where $p_k$ is the single-agent success probability on item $k$ and $\bar p$ its mean \citep{boland1989majority,ladha1992condorcet}; that is, $\rho$ measures heterogeneity of item difficulty rather than interaction between agents. Such a team can be seen as a ``crowd within'' a single model, whose repeated judgments are more correlated than those of different individuals \citep{vul2008crowd, herzog2009wisdom}, and correlated judges add little information \citep{hogarth1978note}.

\begin{proposition} [Task-type scaling under correlation and bias]
\label{thm:scaling_divergence}
Let agents be conditionally independent given the item, and let $N_{\text{eff}}(N) \le 1/\rho$ be defined as above.

\textbf{Disjunctive scaling regime.} With an oracle verifier, accuracy is $1-\mathbb{E}_k[(1-p_k)^N]$, which increases to the fraction of items with $p_k>0$. Under plurality voting, the failure probability on item $k$ decays exponentially in $N$ if the correct answer is the unique mode of $P_k$ and tends to one if it is not, so that accuracy converges to the fraction $\pi$ of items on which the correct answer is the modal answer, that is, of the \emph{easy} items of \citet{chen2024morecalls}. For binary outcomes, the attainable voting gain satisfies
\begin{equation}
\label{eq:voting_bound}
    |\pi - \bar p| \le 2\,\bar p(1-\bar p)(1-\rho).
\end{equation}
The gain from voting is therefore small when $\rho$ is large, even though the exponential convergence of Eq.~\ref{eq:disjunctive_hoeffding} holds on every item with $p_k>\frac12$.

\textbf{Compensatory scaling regime.} For continuous estimates with systematic bias $b$, variance $\sigma^2$, and equicorrelated errors with correlation $\rho$, the mean squared error of the geometric mean is linear in $1/N_{\text{eff}}$:
\begin{equation}
\label{eq:comp_mse}
    \text{MSE}_{\text{comp}}(N) = b^2 + \frac{\sigma^2}{N_{\text{eff}}(N)} \xrightarrow[N \to \infty]{} b^2 + \sigma^2 \rho .
\end{equation}
With item-level biases $b_k$, the same decomposition reads $\mathrm{MSE}(N)=\mathbb{E}_k[b_k^2]+\mathbb{E}_k[\sigma_k^2]/N$, so the attainable reduction is the within-item share of error, $1-\beta$ with $\beta=\mathbb{E}[b_k^2]/(\mathbb{E}[b_k^2]+\mathbb{E}[\sigma_k^2])$. For example, with $\rho \approx 0.9$ (effective capacity $N_{\text{eff}}\approx 1.1$), averaging can remove at most $\sigma^2(1-\rho)\approx 0.1\,\sigma^2$, which is small relative to a dominant bias term $b^2$. Proofs in Appendix~\ref{sec:proofs}.
\end{proposition}

\section{Evaluation of Disjunctive vs. Compensatory Scaling}
\label{sec:empirical_dichotomy}

\subsection{Experimental Setup} 
To evaluate empirically whether LLM multi-agent team performance follows Proposition~\ref{thm:scaling_divergence}, we generated a corpus of approximately $6.8\times10^{7}$ per-agent generations for homogeneous and heterogeneous teams, evaluating all models on the same grid of team sizes $N \in \{1, 2, 3, 5, 7, 10, 15, 20, 25, 30\}$ and communication rounds (Rounds 1 to 3).

\textbf{Models:} We evaluate 13 open-weight models between 3B and 20B parameters, ordered by parameter count: smollm3-3b, qwen2.5-3b, qwen3-4b, mistral-7b, olmo2-7b, qwen2.5-7b, llama3.1-8b, marin-8b, qwen3-8b, phi4-14b, qwen3-14b, r1-distill-qwen-14b, and gpt-oss-20b. Unless stated otherwise, averages are macro-averages over the 13 models.

\textbf{Benchmark Tasks:}  \emph{Disjunctive tasks:} GSM8K \citep{cobbe2021gsm8k} (grade-school mathematics), GSM-Hard \citep{gao2023pal} (GSM8K problems with larger numbers), MATH-500 \citep{hendrycks2021math} (competition mathematics), MMLU-Hard, a difficult subset of MMLU \citep{hendrycks2020mmlu} (multitask knowledge), and ARC-Challenge \citep{clark2018arc} (science question answering). \emph{Compensatory task:} RealFP Fermi problems \citep{fermi2021realfp}, which require numerical order-of-magnitude estimation across physics, geography, and demographics.

\textbf{Inference, Deliberation, and Metric Computation Protocol.}
All models are decoded at temperature $T=0.4$, with three runs per configuration on the full test sets (Appendix~\ref{sec:protocol}).  The prompts request the final answer \emph{before} a brief rationale (Appendix~\ref{sec:prompts}); this answer-first format isolates the answer distribution of the instruction-tuned model itself rather than that of a test-time reasoning procedure, and makes answers robust to truncation.  Two models, gpt-oss-20b and r1-distill-qwen-14b, reason before the formatted answer regardless of the prompt and thus provide a contrast between the two regimes.  In Round~1, agents answer independently; in Rounds~2 and~3, each agent sees its own and its teammates' previous responses and may revise its answer.  Disjunctive answers are aggregated by plurality voting, and we additionally report pass@$N$; Fermi estimates are aggregated by the geometric mean (Eq.~\ref{eq:compensatory_log_mean}) and scored by the absolute log-error (ALE) and by within-one-order-of-magnitude accuracy. We clipped ALE at $\pm3$ orders of magnitude so that implausible reference labels (Section~\ref{sec:label_noise}) and occasional runaway estimates do not dominate the mean. The conclusions do not depend on this threshold (Appendix Table~\ref{tab:clip_sensitivity}).  For RealFP, we analyse the runs generated with the decomposition prompt (10 of the 13 models; Appendix~\ref{sec:integrity}).  Confidence intervals are 95\% item-bootstrap intervals.

\begin{figure}[t]
\centering
\includegraphics[width=\textwidth]{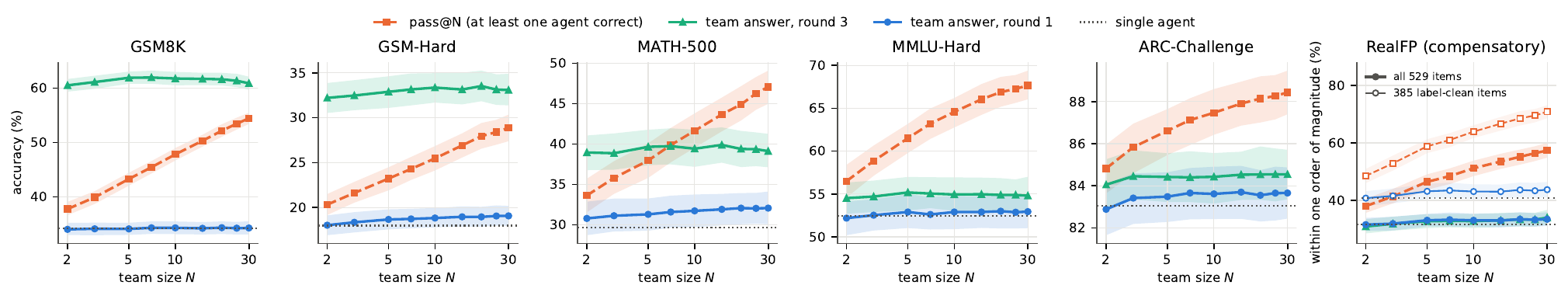}
\caption{Scaling with team size (average over 13 models; bands are 95\% item-bootstrap intervals). The team answer is the plurality vote on the disjunctive tasks and the geometric mean on a representative compensatory  RealFP (right), where accuracy means an estimate within one order of magnitude. Exact values are given in Appendix Table~\ref{tab:task_type_scaling}.}
\label{fig:scaling}
\end{figure}

\subsection{Empirical Scaling Dynamics on Disjunctive vs Compensatory Tasks}
\label{sec:scaling_dynamics}

\textbf{Disjunctive} From Fig. \ref{fig:scaling} and Table~\ref{tab:task_type_scaling} we can see that the probability that at least one of $N$ independent agents is correct (pass@$N$, Section~\ref{sec:disjunctive_model}) rises from the single-agent accuracy to pass@30 of $54.4\%$ on GSM8K (single agent: $34.3\%$), $47.1\%$ on MATH-500 ($29.7\%$), $67.7\%$ on MMLU-Hard ($52.4\%$), $28.9\%$ on GSM-Hard ($18.0\%$), and $88.4\%$ on ARC-Challenge ($83.1\%$). In Steiner's terms, the potential productivity of the team increases with its size. Allowing agents to revise their answers over multiple rounds leads to large accuracy gains on arithmetic tasks. GSM8K accuracy nearly doubles from $34.1\%$ (Round~1, $N=5$) to $61.9\%$ (Round~3, $N=5$); GSM-Hard accuracy rises from $18.7\%$ to $32.9\%$ and MATH-500 accuracy from $31.3\%$ to $39.7\%$. The gain is, however, nearly independent of team size: on GSM8K it is $26.5$ points with a single peer ($N=2$) and $26.6$ points with 29 peers (paired difference $-0.1$, 95\% CI $[-0.6, 0.3]$).  Because the Round-1 format asks for the answer before the rationale, a substantial part of this gain is likely due to the opportunity to reason before committing to an answer; a single agent given a five-fold token budget improves by $12.4$ points on GSM8K (Appendix~\ref{sec:traces} shows an example).

On \textbf{compensatory}  problems, of which Fermi are representative, Fig. \ref{fig:scaling} and Table~\ref{tab:task_type_scaling} show that increasing team size $N$ provides little improvement in either round. Under Round~1 averaging, accuracy within one order of magnitude rises from $31.6\%$ for a single agent to $33.1\%$ at $N=7$ and $33.4\%$ at $N=30$, and the clipped log-error decreases from $1.84$ to $1.73$ at $N=5$ and $1.71$ at $N=30$. Revision does not help: Round-3 accuracy differs from Round-1 accuracy by between $-0.6$ and $+0.7$ points across team sizes. The potential of Fermi teams nevertheless grows: the proportion of items on which at least one agent is within one order of magnitude rises from $31.6\%$ for a single agent to $57.3\%$ at $N=30$ (Section~\ref{sec:basic_model}; Appendix Table~\ref{tab:task_type_scaling}), but averaging cannot select that estimate. Continuous estimates lack verifiable intermediate steps, so revision rarely corrects erroneous orders of magnitude. \label{sec:label_noise}
RealFP contains implausible reference values, such as $2\times10^{90}$ for the number of a person's ancestors; on 144 of the 529 items (27.2\%), the median estimate of the ten models deviates from the reference by more than three orders of magnitude. On the remaining 385 items (open markers in Fig.~\ref{fig:scaling}; fermi-clean in Table~\ref{tab:task_type_scaling}), all levels are higher, but the pattern is unchanged: Round-1 accuracy rises only from $40.9\%$ for a single agent to $43.6\%$ at $N=30$ and the clipped log-error decreases from $1.52$ to $1.39$ at $N=5$ and $1.37$ at $N=30$, whereas pass@30 reaches $70.8\%$. The gap between the potential and the realised performance of Fermi teams therefore does not arise from erroneous labels.

\subsection{The Anatomy of the Disjunctive Ceiling}
\label{sec:disjunctive_ceiling}
  While Section~\ref{sec:scaling_dynamics} showed that disjunctive tasks benefit substantially from multi-agent deliberation compared to compensatory tasks, a critical question remains: \emph{does disjunctive performance scale monotonically as team size grows?} Figure~\ref{fig:scaling} shows that it does not: rather than approaching the pass@$N$ curve, realised disjunctive accuracy exhibits its own ceiling.  In this section, we analyse the empirical evidence and the mechanisms governing this ceiling.

\textbf{The Early Saturation of Independent Ensembles (Round 1)}
\label{sec:disjunctive_independent_ceiling}
Round-1 plurality voting yields almost no improvement with team size: from $N=2$ to $N=30$, accuracy changes by $+0.3$ points on GSM8K (95\% CI $[0.0, 0.5]$), $+1.0$ on GSM-Hard, $+1.3$ on MATH-500, $+0.8$ on MMLU-Hard, and $+0.8$ on ARC-Challenge.  As Proposition~\ref{thm:scaling_divergence} predicts, plurality voting converges to the model's modal answer on each item.  We estimated the item-level answer distributions from the Round-1 answers of the $N=30$ teams (up to 90 samples per item) and predicted plurality accuracy at every team size by resampling.  Across 650 model--task--team-size configurations, predicted and observed accuracies differ by 0.48 points on average (median 0.18; Pearson $r=0.999$; Appendix Figure~\ref{fig:pred_obs}).  The intra-class correlation of correctness ranges from 0.53 to 0.97 (mean 0.78), and the predicted large-team limit exceeds single-agent accuracy by only 0.4--1.2 points per task.  Consistent with Eq.~\ref{eq:voting_bound}, the gain remains within the bound $2\bar p(1-\bar p)(1-\rho)$ for all 65 model--task pairs and uses on average only 11--21\% of it (Appendix Figure~\ref{fig:eq3}).  The two models that reason before answering have a lower $\rho$ ($0.66$ against $0.81$), a larger predicted voting gain ($+3.4$ against $+0.2$ points), and observed gains of $+7.2$ and $+2.8$ points from one to 30 agents, against $-1.0$ to $+1.0$ for the others, consistent with Eq.~\ref{eq:voting_bound} and with the success of self-consistency under chain-of-thought prompting \citep{wang2022self}; the model predicts both regimes (mean error 0.19 and 0.43 points, excluding MATH-500).  A higher temperature is no substitute: for qwen2.5-7b, raising $T$ from 0.2 to 1.0 lowers $\rho$ and raises pass@30 but leaves plurality accuracy unchanged, because the extra samples spread over incorrect answers (Appendix Table~\ref{tab:temp_sweep_disj}).  Additional queries cannot produce a correct plurality on items whose modal answer is wrong; the difference between pass@30 and plurality accuracy at $N=30$ (4.8--20.1 points) is the process loss of plurality voting.

\textbf{The Peak-and-Decay Regime in Deliberated Teams (Round 3)}
\label{sec:disjunctive_deliberation_decay}
Multi-turn communication nearly doubles accuracy on gsm8k ($34.1\% \to 62.0\%$ at the peak) and gsmhard ($18.0\% \to 33.5\%$).  Deliberated teams do not, however, continue to improve with size (Appendix Table~\ref{tab:disjunctive_ceiling_stats}): accuracy peaks at an intermediate team size and declines by $0.4$--$1.1$ points towards $N=30$ on the mathematical and knowledge tasks, most clearly on gsm8k ($-1.0$ points relative to $N=5$, 95\% CI $[-1.4, -0.7]$), and is flat on arc.  The largest declines occur on multi-step symbolic tasks, consistent with an incorrect intermediate result shared by several peers acting as an attractor during revision.

\begin{table}[t]
\caption{\textbf{Per-Model Scaling and Correlation Breakdown on Disjunctive Tasks} (Round~3; accuracy in \%, macro-averaged over the five benchmarks). $\Delta_5 = \text{Acc}_5 - \text{Solo}$ and $\Delta_{\text{decay}} = \text{Acc}_{30} - \text{Acc}_5$, with paired 95\% item-bootstrap intervals in brackets; $\rho$ is the intra-class correlation of correctness. $^\dagger$Best five-model pool, selected and evaluated on disjoint halves of the items (mean over 10 splits; brackets give the range across splits); for pools, $\rho$ is the mean correlation between members.}
\label{tab:per_model_disjunctive}
\begin{center}
\footnotesize
\setlength{\tabcolsep}{2.0pt}
\resizebox{0.9\textwidth}{!}{
\begin{tabular}{lcccccccc}
\toprule
\textbf{Model} & \textbf{Solo} & $\mathbf{N=5}$ & $\mathbf{N=30}$ & $\mathbf{\Delta_5}$ & $\mathbf{\Delta_{\text{decay}}}$ & $\bm{\rho}$ & $\mathbf{N_{\text{eff}}}$ & $\mathbf{N_{\text{eff}}}$ \\
 & \textbf{Acc} & \textbf{Acc} & \textbf{Acc} & \textbf{Gain} & & & \textbf{(5)} & \textbf{(30)} \\
\midrule
\multicolumn{9}{l}{\textbf{Homogeneous Architectures (Multi-Turn Deliberation, Round~3)}} \\
smollm3-3b & 32.43 & 45.09 & 45.61 & $+12.66$\,{\scriptsize[+11.80, +13.55]} & $+0.52$\,{\scriptsize[-0.11, +1.15]} & 0.67 & 1.36 & 1.47 \\
qwen2.5-3b & 27.08 & 32.32 & 32.58 & $+5.23$\,{\scriptsize[+4.57, +5.88]} & $+0.26$\,{\scriptsize[-0.31, +0.80]} & 0.82 & 1.16 & 1.20 \\
qwen3-4b & 42.13 & 51.21 & 52.71 & $+9.09$\,{\scriptsize[+8.34, +9.87]} & $+1.50$\,{\scriptsize[+1.05, +1.97]} & 0.86 & 1.13 & 1.16 \\
mistral-7b & 26.14 & 28.96 & 27.34 & $+2.82$\,{\scriptsize[+2.34, +3.30]} & $-1.62$\,{\scriptsize[-2.09, -1.19]} & 0.84 & 1.15 & 1.18 \\
olmo2-7b & 27.17 & 41.73 & 39.26 & $+14.56$\,{\scriptsize[+13.54, +15.46]} & $-2.47$\,{\scriptsize[-3.16, -1.77]} & 0.74 & 1.26 & 1.34 \\
qwen2.5-7b & 40.49 & 53.61 & 52.29 & $+13.12$\,{\scriptsize[+12.12, +14.06]} & $-1.32$\,{\scriptsize[-1.88, -0.78]} & 0.91 & 1.08 & 1.10 \\
llama3.1-8b & 32.47 & 50.22 & 49.44 & $+17.75$\,{\scriptsize[+16.80, +18.71]} & $-0.78$\,{\scriptsize[-1.53, -0.01]} & 0.77 & 1.23 & 1.29 \\
marin-8b & 28.62 & 45.75 & 42.74 & $+17.14$\,{\scriptsize[+16.19, +18.03]} & $-3.01$\,{\scriptsize[-3.71, -2.33]} & 0.64 & 1.40 & 1.53 \\
qwen3-8b & 51.02 & 61.90 & 63.21 & $+10.89$\,{\scriptsize[+10.06, +11.67]} & $+1.31$\,{\scriptsize[+0.77, +1.85]} & 0.81 & 1.18 & 1.22 \\
phi4-14b & 48.40 & 69.30 & 68.20 & $+20.90$\,{\scriptsize[+19.86, +21.99]} & $-1.10$\,{\scriptsize[-1.66, -0.49]} & 0.87 & 1.12 & 1.15 \\
qwen3-14b & 50.19 & 63.37 & 63.71 & $+13.18$\,{\scriptsize[+12.28, +14.05]} & $+0.34$\,{\scriptsize[-0.08, +0.79]} & 0.93 & 1.06 & 1.07 \\
r1-distill- & 74.50 & 81.17 & 83.01 & $+6.68$\,{\scriptsize[+5.64, +7.64]} & $+1.84$\,{\scriptsize[+1.13, +2.53]} & 0.64 & 1.40 & 1.53 \\
gpt-oss-20b & 84.72 & 87.87 & 88.24 & $+3.16$\,{\scriptsize[+2.62, +3.71]} & $+0.37$\,{\scriptsize[-0.00, +0.75]} & 0.68 & 1.34 & 1.45 \\
\midrule
\multicolumn{9}{l}{\textbf{Heterogeneous Combinations (Single-Agent Answers, Majority Voting)}} \\
Hetero Best$^\dagger$ & 61.02 & 83.16 & --- & $+22.14$\,{\scriptsize[+21.36, +22.74]$^\dagger$} & --- & 0.37 & 2.01 & --- \\
Hetero 7B--8B & 31.93 & 39.76 & --- & $+7.83$\,{\scriptsize[+7.02, +8.52]} & --- & 0.54 & 1.58 & --- \\
\bottomrule
\end{tabular}
}
\end{center}
\end{table}

 \textbf{Per-Model Scaling and Correlation Breakdown}
\label{sec:per_model_disjunctive}
Deliberation gains at $N=5$ range from $+2.8$ (mistral-7b) to $+20.9$ points (phi4-14b; Table~\ref{tab:per_model_disjunctive}) and are not explained by $\rho$ (rank correlation $0.12$, $p=0.71$).  The two strongest models, gpt-oss-20b and r1-distill-qwen-14b, gain little ($+3.2$ and $+6.7$ points), since they reason before the formatted answer.

 \textbf{Asymptotic Capacity Saturation ($N_{\text{eff}} \le 1/\rho$) and Post-Peak Decline ($\Delta_{\text{decay}} < 0$):} With $\rho$ between 0.64 and 0.93, $N_{\text{eff}}(5)$ lies between 1.06 and 1.40 and $N_{\text{eff}}(30)$ between 1.07 and 1.53, so a six-fold larger team adds almost no effective capacity. Scaling from $N=5$ to $N=30$ significantly reduces Round-3 accuracy in 6 of the 13 models (by up to $-3.01$ points for marin-8b, 95\% CI $[-3.71, -2.33]$, and up to $-6.1$ points on GSM8K), significantly increases it in three, and leaves it unchanged in four.

 \textbf{On Architectural Diversity}
The last rows of Table~\ref{tab:per_model_disjunctive} evaluate majority voting over the single-agent answers of five different models, choosing the best of the $\binom{8}{5} = 56$ combinations of the eight strongest panel models on a random half of the items and evaluating it on the other half.  The most frequently selected pool (gpt-oss-20b, r1-distill-qwen-14b, qwen3-8b, phi4-14b, smollm3-3b) has a pairwise error correlation of $\rho = 0.37$ ($N_{\text{eff}}(5) = 2.01$) and reaches $83.2\%$ on held-out items, $+22.1$ points above the mean of its members, but it falls short of its strongest member (gpt-oss-20b, $86.8\%$) in all 10 splits, by 3.4 to 4.3 points, consistent with \citet{pappu2026experts}; equal-weight voting is suboptimal when members differ in competence \citep{nitzan1982optimal}. Furthermore, five 7B--8B models from different developers (qwen2.5-7b , llama3.1-8b , mistral-7b , olmo2-7b , marin-8b) have $\rho = 0.54$ ($N_{\text{eff}} = 1.58$) and reach $39.8\%$, $+7.8$ points above the mean member (95\% CI $[7.0, 8.5]$) but $2.5$ points below the strongest member, qwen2.5-7b ($[1.2, 3.8]$).  A real heterogeneous run with the same models (100 items per task) confirms this ($22.2\%$ against $30.3\%$ for qwen2.5-7b at $N=5$).

\subsection{The Anatomy of the Compensatory Ceiling}
\label{sec:ceiling_diagnosis}
\label{sec:compensatory_ceiling}

Section~\ref{sec:scaling_dynamics} showed that compensatory tasks benefit little from multi-agent scaling or deliberation.  Table~\ref{tab:per_model_fermi} provides a per-model breakdown.  For homogeneous models, the error reduction from a single agent to a team of five ranges from $0.9\%$ to $8.1\%$, with one exception (marin-8b, $15.4\%$), and increasing the team to $N=30$ yields little further reduction.  The intra-class correlation of signed log-errors lies between 0.70 and 0.95, and effective team capacity $N_{\text{eff}}(5)$ lies between $1.04$ and $1.32$: five agents provide the variance reduction of little more than one.  Averaged over models, item-level bias accounts for $\beta = 0.87$ of the mean squared log-error (between-item standard deviation of $b_k$: 1.96 orders of magnitude; within-item standard deviation: 0.75), so that even an infinite homogeneous team could remove at most about 13\% of it.  Eq.~\ref{eq:comp_mse} predicts the observed error of the geometric mean at every team size to within 0.6\% on average for all ten models (Pearson $r=0.997$; Appendix Figure~\ref{fig:eq4}).  Only the model with the most dispersed samples (marin-8b) and heterogeneous multi-family teams depart noticeably from the ceiling: the 7B--8B pool reduces the error of its average member by $25.8\%$ (95\% CI [23.1, 28.7]) and, with $1.45$, is also more accurate than its best member alone ($1.80$). \textbf{Temperature Invariance} For four models at $T \in \{0.4, 0.8, 1.0\}$ (Appendix Table~\ref{tab:temperature_grid}), Fermi accuracy is virtually identical across temperatures: higher temperature increases the within-item dispersion of log-errors by 30--70\%, but the intra-class correlation remains between 0.83 and 0.98.

\begin{table}[h] 
\caption{\textbf{Per-Model Scaling Breakdown on Compensatory Fermi Estimation} (RealFP, 529 items, Round~1, decomposition-prompt runs). MAE is the mean absolute log-error of the geometric mean (clipped at $\pm3$); ``Drop'' is its relative reduction from $N=1$ to $N=5$, with a paired 95\% item-bootstrap interval; $\rho$ is the intra-class correlation of signed log-errors. $^\ddagger$The five 7B--8B models of Section~\ref{sec:disjunctive_ceiling}, one answer each ($N=5$) or six each ($N=30$); ``Solo'' is their mean, and $\rho$ their mean cross-model bias correlation.}
\label{tab:per_model_fermi}
\begin{center}
\footnotesize
\setlength{\tabcolsep}{2.5pt}
\resizebox{0.9\textwidth}{!}{
\begin{tabular}{lcccccccc}
\toprule
\textbf{Model} & \textbf{Solo} & $\mathbf{N=5}$ & $\mathbf{N=30}$ & \textbf{Drop} & \textbf{Solo} & $\mathbf{N=5}$ & $\bm{\rho}$ & $\mathbf{N_{\text{eff}}}$ \\
 & \textbf{MAE} & \textbf{MAE} & \textbf{MAE} & \textbf{\%} & \textbf{Acc} & \textbf{Acc} & & \textbf{(5)} \\
\midrule
smollm3-3b & 1.85 & 1.71 & 1.67 & $+7.3$\,{\scriptsize[+4.7, +9.8]} & 31.9 & 33.1\,{\scriptsize[29.4, 36.8]} & 0.81 & 1.18 \\
qwen2.5-3b & 1.94 & 1.86 & 1.84 & $+4.1$\,{\scriptsize[+2.0, +6.1]} & 28.6 & 26.1\,{\scriptsize[22.9, 29.5]} & 0.87 & 1.12 \\
mistral-7b & 2.08 & 1.91 & 1.91 & $+8.1$\,{\scriptsize[+6.1, +11.0]} & 23.4 & 27.7\,{\scriptsize[23.6, 31.6]} & 0.87 & 1.12 \\
olmo2-7b & 1.80 & 1.70 & 1.68 & $+5.5$\,{\scriptsize[+3.1, +7.3]} & 34.1 & 34.5\,{\scriptsize[31.3, 38.7]} & 0.87 & 1.11 \\
qwen2.5-7b & 1.80 & 1.72 & 1.71 & $+3.9$\,{\scriptsize[+2.3, +5.6]} & 34.0 & 33.9\,{\scriptsize[30.1, 37.7]} & 0.90 & 1.08 \\
llama3.1-8b & 2.02 & 1.87 & 1.84 & $+7.9$\,{\scriptsize[+5.0, +10.7]} & 23.7 & 26.7\,{\scriptsize[23.1, 30.2]} & 0.84 & 1.15 \\
marin-8b & 2.07 & 1.75 & 1.69 & $+15.4$\,{\scriptsize[+13.0, +17.7]} & 24.1 & 31.2\,{\scriptsize[27.9, 34.6]} & 0.70 & 1.32 \\
qwen3-8b & 1.64 & 1.61 & 1.60 & $+2.2$\,{\scriptsize[+0.2, +4.8]} & 38.7 & 38.4\,{\scriptsize[33.8, 42.5]} & 0.95 & 1.04 \\
phi4-14b & 1.66 & 1.61 & 1.61 & $+2.7$\,{\scriptsize[+0.6, +5.0]} & 39.2 & 41.0\,{\scriptsize[37.1, 45.6]} & 0.94 & 1.05 \\
qwen3-14b & 1.59 & 1.57 & 1.57 & $+0.9$\,{\scriptsize[-1.6, +3.1]} & 38.4 & 38.6\,{\scriptsize[34.4, 42.7]} & 0.93 & 1.06 \\
\midrule
\multicolumn{9}{l}{\textbf{Heterogeneous:}} \\
Hetero 7B--8B$^\ddagger$ & 1.95 & 1.45 & 1.41 & $+25.8$\,{\scriptsize[+23.1, +28.7]} & 27.8 & 38.4\,{\scriptsize[34.2, 42.3]} & 0.61 & 1.45 \\
\bottomrule
\end{tabular}
}
\end{center}
\end{table}

\textbf{Empirical Analysis of Unbiasedness}
  The classical model in Section~\ref{sec:basic_model} assumes that errors balance out to zero on average ($\mathbb{E}[e_i] = 0$).  To test this condition, we measure the distribution of signed log-errors $e_i = \log_{10}(x_i / g)$ in Appendix Table~\ref{tab:unbiased_stats}.  Models such as mistral-7b ($b = +0.71$), qwen2.5-3b ($b = +0.65$), and llama3.1-8b ($b = +0.62$) overestimate by more than half an order of magnitude on average, and 61--64\% of their estimates exceed the ground truth.  Conversely, marin-8b ($b = -0.20$, 95\% CI $[-0.38, -0.03]$) significantly underestimates.  Even for the models whose global mean is not significantly different from zero (qwen2.5-7b, $b = +0.11$, $[-0.05, 0.28]$, and qwen3-8b, $b = -0.04$, $[-0.20, 0.13]$), the unbiasedness condition fails at the level of individual items.  For a given prompt $k$, independent samples from the same model cluster around the model's item-specific estimate, so that $\mathbb{E}[e_i \mid \text{item } k] = b_k \neq 0$.  The root-mean-square item-level bias, averaged across models, is $2.01$ orders of magnitude (with clipping at $\pm3$).  Intra-model averaging reduces sampling variation around $b_k$ but leaves the systematic error $b_k$ untouched (Appendix~\ref{sec:traces} shows an example).

\textbf{Cross-Model Error Correlation and Shared Directional Misconceptions.}
If individual models have non-zero item-level biases, can a multi-model ensemble cancel these errors?  That is, does $\sum_{m=1}^M \mathbb{E}[e^{(m)} \mid \text{item } k] \approx 0$?  To answer this question, we compute the pairwise correlation of item-level biases across models (Appendix Table~\ref{tab:cross_model_corr}).  The mean pairwise correlation is $r = 0.63$ (95\% CI [0.60, 0.67]; standard deviation across model pairs 0.06), ranging from $r = 0.46$ between mistral-7b and qwen3-8b to $r = 0.77$ between qwen3-8b and qwen3-14b. On $31.5\%$ of all items, \emph{every model errs in the same direction}.  Because these correlations are well below one, averaging across model families can cancel part of the item-level bias, which is consistent with the gains of the heterogeneous pool in Table~\ref{tab:per_model_fermi}. Part of this alignment is due to erroneous reference labels: on the 385 label-clean items (Section~\ref{sec:label_noise}), the mean correlation falls to $r=0.41$ and all models err in the same direction on $19.1\%$ of items, whereas the bias share remains high ($\beta=0.82$).

\section{Discussion}
\label{sec:discussion}

\textbf{Implications.}
On disjunctive tasks (Section~\ref{sec:disjunctive_ceiling}), the difference between pass@$N$ and plurality accuracy identifies the process loss that verifiers, reward models, or structured deliberation must recover; reasoning before answering or combining model families can raise the voting ceiling, by an amount that Eq.~\ref{eq:voting_bound} bounds.  On compensatory tasks (Section~\ref{sec:compensatory_ceiling}), reducing the item-level bias requires changing the estimates themselves, for instance by combining models whose biases are not aligned, rather than adding samples; prompting interventions and a fine-tuned model are evaluated in Appendix~\ref{sec:interventions}. 

\textbf{Limitations.} First, the prompts request the answer before the rationale, which lowers absolute accuracy and shapes the deliberation gain; chain-of-thought prompting was not evaluated for the answer-first models, so the size of the voting ceiling is specific to this regime, whereas its mechanism is not. Second, there is no revision condition without peers.  Third, the Fermi results rest on one benchmark with substantial label noise, generated with two prompt families (Appendix~\ref{sec:integrity}).  Fourth, all models are open-weight with at most 20B parameters, and additive, conjunctive, and discretionary tasks remain to be studied.

\section*{Reproducibility Statement}
The theoretical results rest on the assumptions stated in Proposition~\ref{thm:scaling_divergence}; the ideal-case analysis and complete proofs are given in Appendix~\ref{sec:ideal_dichotomy} and Appendix~\ref{sec:proofs}.  Models (with Hugging Face identifiers), prompts, decoding parameters, the number of runs per configuration, and the benchmark items used are specified in Appendix~\ref{sec:prompts}, and the data-processing steps, including the identification of the RealFP prompt families and the corrections and exclusions of the log audit, in the data-integrity paragraph of the same appendix.  The metrics, the clipping of log-errors, the label-clean RealFP subset, and the item-bootstrap intervals are defined in Section~\ref{sec:empirical_dichotomy}, with sensitivity analyses in Appendices~\ref{app:disj} and~\ref{app:comp}.  The per-agent generation logs will be released upon publication; the analysis scripts, which reproduce all tables and figures from these logs, will be open sourced upon acceptance.

\section*{Use of Large Language Models}
 The research question, the application of Steiner's taxonomy to multi-agent LLM systems, and the design and execution of the experiments are the authors' own.  We used generative AI tools (a coding and writing assistant) to provide feedback on the methodology in the form of a simulated review; to help correct and refine the theoretical analysis, including the voting-gain bound (Eq.~\ref{eq:voting_bound}), the compensatory error decomposition (Eq.~\ref{eq:comp_mse}), and parts of their proofs; to implement analysis code, including the bootstrap intervals, the validation of Eqs.~\ref{eq:voting_bound} and~\ref{eq:comp_mse}, the identification of prompt families and implausible reference labels, and the sensitivity analyses; and to draft interpretations of the results.  We did not use generative AI tools to generate synthetic data, and qualitative or thematic data analysis is not applicable to this work.  Additionally, we used these tools to create figures, draft and edit parts of the text, suggest relevant literature, and format and check references.  All references, all sections of the paper including text, were revised manually by all authors.  AI-generated code was reviewed by the authors, and all reported numbers were recomputed from the raw generation logs; every suggested reference was checked against its published source. We take responsibility for the final content of this work, including text, claims, and artifacts produced with the aid of generative AI.

\bibliography{references_steiner_cc}
\bibliographystyle{iclr2027_conference}

\appendix

\section{Detailed Steiner Taxonomy Mapping on all 5 task categories}
\label{sec:other_cells}

\paragraph{Disjunctive Tasks}in which  the outcome is established through $Y = \max_i \{y_i\}$ or plurality/majority/verifier selection and/or debate. The tasks are unitary and optimizing and group success requires that \emph{at least one} agent discovers a valid solution, and the group then recognizes it as such. In recent literature, example benchmarks falling under this task type are: i) multi-step mathematical reasoning such as MATH-500 \citep{hendrycks2021math, du2023improving} and GSM8K \citep{cobbe2021gsm8k}, ii) challenging multiple-choice knowledge suites such as MMLU-Pro \citep{wang2024mmlu} and GPQA Diamond \citep{rein2024gpqa}, iii) school science answering in ARC-Challenge \citep{clark2018arc}, and iv) functional code generation in BigCodeBench \citep{zhuo2024bigcodebench}. Multi-agent LLM systems operationalize this task structure via plurality voting \citep{wang2022self}, verifier re-ranking, and multi-agent debate \citep{du2023improving, wang2025mixture}, in some studies reporting success while in others showing diminishing benefits with scale.

\paragraph{Compensatory Tasks}in which the outcome is determined by $Y = \text{Aggr}(x_1, \dots, x_n)$ via  statistical aggregation (geometric mean, arithmetic mean, or median) of continuous numerical estimates ($x_i \in \mathbb{R}^+$) or calibrated probabilities ($p_i \in [0, 1]$).  In recent AI research, this type is exemplified by post-cutoff future event forecasting in ForecastBench \citep{forecastbench2024}, numerical guesstimation of world-model properties in Guesstimate \citep{chuang2025probing}, multi-scale physical and dimensional estimation in RealFP and SynthFP \citep{fermi2021realfp}, and competitive Science Olympiad estimation in SciOly. LLMs typically perform worse on such tasks types compared to disjunctive types \cite{liu2025fermi,epstein2025fermieval}.

\paragraph{Additive Tasks} in which the outcome is established through $Y = \sum_i y_i$ or $Y = \bigcup_i S_i$ via  cumulative sum, total coverage, or deduplicated set union of independent member efforts. These tasks are divisible and maximizing by nature. This type represents the classical study by \citet{ingham1974ringelmann} and \citet{latane1979many}, where social loafing and coordination decay originally manifested. And this is where the strongest member determines the overall performance of the team. Existing benchmarks that may fall under this category include open-ended creative brainstorming and hypothesis generation \citep{olson2021naming}, divergent semantic exploration in the Divergent Association Task (DAT), and parallel synthetic instruction generation. 

\paragraph{Conjunctive Tasks} in which the outcome is established through $Y = \min_i \{y_i\}$ via weakest-link coupling). Group success requires that \emph{all} agents or subtasks succeed without failure; an error at any intermediate stage can invalidate the entire collective output. Looking at  real-world repository-level software engineering in SWE-bench and SWE-bench Verified \citep{jimenez2024swebench}, end-to-end multi-agent enterprise workflows in TheAgentCompany \citep{theagentcompany2025}, interactive multi-turn operating system manipulation in OSWorld \citep{xie2024osworld}, and sequential tool-calling environments in WorkBench and BrowseComp-Plus \citep{kim2025towards}, we notice that they can be solved as a sequence of sub-tasks where a failure in any intermediate step causes zero task completion.

\paragraph{Discretionary Tasks} in which the outcome is established through $Y = f_{\text{group}}(\{y_i\})$ via group discretion.
In Steiner's taxonomy, discretionary tasks grant the group complete autonomy to select its own combination rule; the task constraints themselves do not mandate how member contributions combine. Modern multi-agent meta-orchestration frameworks (e.g., router-judge architectures, Mixture-of-Agents \citep{wang2025mixture}, ChatEval \citep{chan2023chateval}, and AgentBench \citep{liu2023agentbench}) where a central orchestrator dynamically decides whether to average, vote, deliberate, or delegate to specialized tool-calling agents may correspond to this category. 

\FloatBarrier

\section{Theoretical Details}
\label{app:theory}

\subsection{The Scaling Dichotomy Under Ideal Assumptions}
\label{sec:ideal_dichotomy}

  Comparing the ideal disjunctive model (Section~\ref{sec:disjunctive_model}) and the ideal compensatory model (Section~\ref{sec:basic_model}) clarifies what differs between the two task types.

\textbf{Disjunctive Tasks (Selection / Voting):} The candidate output space $\mathcal{Y}$ is discrete. Plurality voting and verifier selection isolate the correct modal response from dispersed errors. Under Condorcet's jury theorem and Hoeffding's bound, the collective failure probability decays exponentially:
    \begin{equation}
        P(\text{Disjunctive Error}) \le \exp\left( -2N \left( p - \frac{1}{2} \right)^2 \right) = \mathcal{O}(e^{-\alpha N}).
    \end{equation}

\textbf{Compensatory Tasks (Statistical Aggregation):} Estimates $x_i \in \mathbb{R}^+$ and signed errors $e_i = \log_{10}(x_i / g) \in \mathbb{R}$ are continuous, and the collective relies on statistical cancellation. The standard error decays as
    \begin{equation}
        \text{SE}(\bar{e}_N) = \frac{\sigma}{\sqrt{N}} = \mathcal{O}(N^{-1/2}),
    \end{equation}
    and, for sub-Gaussian errors, the probability that $|\bar e_N|$ exceeds a fixed tolerance $\epsilon$ also decays exponentially, as $\exp(-N\epsilon^2/(2\sigma^2))$.

  Under ideal independence and unbiasedness, both task types therefore enjoy exponential suppression of the probability of a large error.  The difference between them lies in how they respond to correlation and bias, which we analyse next.

\subsection{Correlated and Biased LLM Ensembles: Breakdown Across Task Types}
\label{sec:llm_breakdown}

  The classical models in Sections~\ref{sec:disjunctive_model} and~\ref{sec:basic_model} assume independent and unbiased members.  Ensembles of samples from the same LLM violate both assumptions at the level of individual items, and the consequences differ across Steiner's task classes.

\subsubsection{Disjunctive Tasks Under Correlation and Bias}

In discrete reasoning tasks (e.g., GSM8K, MATH-500, MMLU), agents exhibit positive pairwise correlation $\rho > 0$:

\textbf{Correlated Condorcet Voting:} When individual votes are exchangeable with pairwise correlation $\rho > 0$, the variance of the collective vote count is inflated by the factor $1 + (N-1)\rho$ \citep{boland1989majority, ladha1992condorcet}, which corresponds to the effective size $N_{\text{eff}}(N) = N/(1 + (N-1)\rho)$. For independent samples from one model, this correlation arises from heterogeneous item difficulty, and plurality accuracy converges to the fraction of items on which the correct answer is modal (Proposition~\ref{thm:scaling_divergence}).

\textbf{Error Dispersion in Generative Search:} In open-ended reasoning, the space of incorrect derivations is large. When incorrect answers are dispersed across many values while correct derivations converge on $y^*$, the correct answer can remain the plurality even for $p_k<\frac12$, which allows voting to succeed despite substantial correlation. When a model concentrates its probability on a single incorrect answer, however, additional samples reinforce that answer; our measurements show that this concentration is common, with intra-class correlations between 0.53 and 0.97.

\textbf{Verifiable Truth Propagation in Deliberation:} Mathematical reasoning steps can be checked, that is, correct answers are demonstrable \citep{laughlin1986demonstrability}. When agents exchange rationales, an agent with a correct step can present an argument that its peers are able to verify.

\subsubsection{Compensatory Tasks Under Correlation and Bias}
  Continuous numerical estimation (e.g., Fermi estimation in RealFP) offers fewer structural defences against correlated errors:

\textbf{Violation of Unbiasedness ($b \neq 0$):} LLMs internalise miscalibrated priors about physical, demographic, and geographical quantities, producing item-level biases $\mathbb{E}[e_i \mid k] = b_k \neq 0$. Because geometric averaging is linear in log space, the expected team error is invariant to ensemble size:
    \begin{equation}
        \mathbb{E}[\bar{e}_N \mid k] = \frac{1}{N}\sum_{i=1}^N \mathbb{E}[e_i \mid k] = b_k \implies \text{Bias}^2(\bar{e}_N \mid k) = b_k^2 \quad \text{for all } N.
    \end{equation}
    Averaging samples of one model cannot reduce its item-level bias.

\textbf{Violation of Independence ($\rho > 0$):} With equicorrelated errors, $\text{Cov}(e_i, e_j) = \rho \sigma^2 > 0$, and the variance of the average approaches a non-zero floor:
    \begin{equation}
        \text{Var}(\bar{e}_N) = \frac{\sigma^2}{N} + \frac{N-1}{N}\rho \sigma^2 \xrightarrow[N \to \infty]{} \rho \sigma^2 .
    \end{equation}
    Empirically, the intra-class correlation of signed log-errors lies between 0.70 and 0.95 (Table~\ref{tab:per_model_fermi}).

\textbf{Limited Verifiability in Deliberation:} Order-of-magnitude estimates lack verifiable intermediate steps in zero-shot dialogue. Consistent with this, revision leaves Fermi accuracy essentially unchanged (between $-0.6$ and $+0.7$ points).

\subsection{Proofs}
\label{sec:proofs}

\paragraph{Oracle accuracy.}  Given item $k$, the $N$ outputs are independent, so the probability that none is correct is $(1-p_k)^N$; averaging over items and applying monotone convergence gives the limit $\Pr_k(p_k>0)$.

\paragraph{ Plurality voting.}  This argument generalises Lemma~1 of \citet{chen2024morecalls}, whose difficulty indicator $d_V(x)$ equals $-\delta_k$, to arbitrary answer sets.  Fix item $k$ and a competing answer $y\neq y_k^\ast$.  The variables $D_i=\mathbb{1}[Y_{k,i}=y_k^\ast]-\mathbb{1}[Y_{k,i}=y]\in[-1,1]$ are independent with mean $\delta_{k,y}=p_k-P_k(y)$.  If $\delta_{k,y}>0$, Hoeffding's inequality gives $\Pr(\sum_i D_i\le0)\le\exp(-N\delta_{k,y}^2/2)$, and a union bound over competitors yields exponential convergence to success; if $\delta_{k,y}<0$ for some $y$, the same argument shows that $y_k^\ast$ loses with probability tending to one.  Averaging over items yields the limit $\pi$.

\paragraph{Attainable voting gain.}  For binary outcomes, item $k$ is decided correctly in the limit if and only if $p_k>\frac12$, so $|\pi-\bar p|\le\mathbb{E}_k[\min(p_k,1-p_k)]\le2\,\mathbb{E}_k[p_k(1-p_k)]$, since $\min(p,1-p)\le2p(1-p)$ on $[0,1]$.  Finally, $\mathbb{E}_k[p_k(1-p_k)]=\bar p(1-\bar p)-\mathrm{Var}_k(p_k)=\bar p(1-\bar p)(1-\rho)$.

\paragraph{Compensatory decomposition.}  With $e_{k,i}=b_k+\varepsilon_{k,i}$ and conditionally independent $\varepsilon_{k,i}$ of variance $\sigma_k^2$, $\mathbb{E}[\bar e_{k,N}^2\mid k]=b_k^2+\sigma_k^2/N$.  For equicorrelated errors with common bias $b$, $\mathrm{Var}(\bar e_N)=\sigma^2(1+(N-1)\rho)/N=\sigma^2/N_{\text{eff}}(N)$, which gives the stated MSE.

\FloatBarrier

\section{Experimental Details: Models, Prompts, and Data Integrity}
\label{sec:prompts}

\paragraph{Evaluation protocol.}
\label{sec:protocol}
Models are served with vLLM with nucleus parameter $0.95$ and at most 1{,}024 generated tokens; explicit thinking modes are disabled, and gpt-oss-20b uses low reasoning effort.  Each configuration is run with three random seeds (two for qwen2.5-3b, one or two for r1-distill-qwen-14b) on the benchmark test sets: 1{,}319 items for GSM8K, 1{,}017 for GSM-Hard, 500 for MATH-500, 724 for MMLU-Hard, 1{,}165 for ARC-Challenge, and the 529 of 557 RealFP test items with a positive numerical answer.  In Rounds~2 and~3, each agent receives its own previous response and the previous responses of all $N-1$ teammates, each truncated to 120 tokens (answer, confidence, and the beginning of the rationale).  Plurality ties are broken at random.  On disjunctive tasks, $\rho$ is the intra-class correlation of binary correctness indicators across items, macro-averaged over the five benchmarks, and $N_{\text{eff}}(N) = N/(1 + (N-1)\rho)$.  On RealFP, the item-level bias $b_k$ is the mean signed log-error of a model on item $k$, $\rho$ is the intra-class correlation of signed log-errors, and an estimate is counted as accurate if $|\bar{e}_{N}| \le 1$.  Confidence intervals are percentile intervals from 1{,}000--2{,}000 bootstrap resamples of items, with items resampled jointly across models and differences computed on paired items.

 {\sloppy\paragraph{Models.} Hugging Face identifiers, as recorded in the run manifests: \path{HuggingFaceTB/SmolLM3-3B}, \path{Qwen/Qwen2.5-3B-Instruct}, \path{Qwen/Qwen3-4B}, \path{mistralai/Mistral-7B-Instruct-v0.3}, \path{allenai/OLMo-2-1124-7B-Instruct}, \path{Qwen/Qwen2.5-7B-Instruct}, \path{meta-llama/Llama-3.1-8B-Instruct}, \path{marin-community/marin-8b-instruct}, \path{Qwen/Qwen3-8B}, \path{microsoft/phi-4}, \path{Qwen/Qwen3-14B}, \path{deepseek-ai/DeepSeek-R1-Distill-Qwen-14B}, and \path{openai/gpt-oss-20b}.\par}

\paragraph{Prompts.}  The Round-1 prompt for GSM8K reads: ``Solve this math problem step by step.  Be concise.  Problem: \{question\}.  Format EXACTLY (answer and confidence FIRST, then reasoning): FINAL: $\langle$integer$\rangle$ CONF: $\langle$0--100$\rangle$ RATIONALE: $\langle$brief reasoning, max 4 lines$\rangle$.'' The other tasks use the same structure with task-specific answer formats.  The revision prompt presents the agent's previous response and the peer responses, each truncated to 120 tokens, and asks the agent to ``update or keep'' its answer in the same format.  For RealFP, the prompt analysed in the main text instructs the model to decompose the quantity into a chain of factors with units, to combine them while tracking units and powers of ten, and to answer in the unit of the reference value; its revision prompt states that ``your goal is NOT consensus'' and asks the agent to change its estimate only to fix a concrete error in its own work.  An earlier Fermi prompt, used for some runs (see below), was derived from the GSM8K template (``Give your best order-of-magnitude estimate for this Fermi problem.  Reason step by step.'', with a rationale of at most four lines) and used the same revision prompt as the disjunctive tasks.

\paragraph{Data integrity.}
\label{sec:integrity}
  An audit of the generation logs identified four irregularities, all of which are corrected or excluded in the reported analyses.  (i) The RealFP runs were generated with the two prompt families described above, which we identify from the agents' outputs: in every run of the decomposition family, 14--99\% of Round-2 rationales echo the diversity-preserving instruction (e.g., ``I kept my estimate because \dots''), and most Round-1 outputs list numbered factors, whereas no Round-2 rationale of the earlier family does.  The decomposition family covers ten of the 13 panel models (all runs of qwen2.5-3b , olmo2-7b , qwen2.5-7b, and marin-8b; the third run of mistral-7b , llama3.1-8b , qwen3-8b , phi4-14b, and qwen3-14b; and the second and third runs of smollm3-3b), and all Fermi results in the main text are restricted to these runs.  gpt-oss-20b , qwen3-4b, and r1-distill-qwen-14b were run only with the earlier prompt.  For the earlier family (nine models), the clipped log-error changes from 1.83 ($N=1$) to 1.72 ($N=5$) and 1.74 ($N=30$), the bias share is $\beta=0.89$, and the mean cross-model bias correlation is 0.55 (0.33 on label-clean items), so the conclusions are unchanged.  (ii) One cell of the earlier family (gpt-oss-20b, RealFP, $N=30$, third run) has 1.4 times longer outputs and half the accuracy of all other cells and is excluded.  (iii) r1-distill-qwen-14b frequently reaches the 1{,}024-token limit: single-agent outputs are truncated on 44\% of MATH-500, 31\% of GSM-Hard, 29\% of MMLU-Hard, and 3\% of GSM8K and ARC-Challenge items, and on RealFP at $N\ge10$; its results on these tasks should be read as lower bounds.  (iv) The original answer parser did not recognise exponents written with superscript digits and retained only the mantissa; this affected 16.0\% of the Fermi answers of qwen3-14b, 5.7\% of those of gpt-oss-20b, and at most 1\% of those of every other model, and all Fermi answers were re-parsed before analysis.

\FloatBarrier

\subsection{Example Traces}
\label{sec:traces}

  To illustrate the mechanisms discussed in Sections~\ref{sec:scaling_dynamics} and~\ref{sec:compensatory_ceiling}, we show one disjunctive and one compensatory trace.  To avoid selective reporting, each example was drawn at random (fixed seed) from all items that satisfy a pre-specified rule; each example is the first draw under its rule.  Outputs are abbreviated with ellipses but otherwise reproduced verbatim.

\paragraph{Disjunctive example (llama3.1-8b, GSM8K, $N=2$, first run).}  Selection rule: both agents are incorrect in Round~1 and the plurality team answer is correct in Round~3 (595 items).
  \begin{quote}\small
\textbf{Question:} Carson lives in an old house where the pipes will freeze if the temperature inside the house falls below 32 degrees. Every log of wood Carson burns heats the house up by 5 degrees. If it was 45 degrees during the day, and it's 33 degrees colder during the night compared to the day, how many logs does Carson need to burn to prevent the pipes from freezing? (Reference: 4.)\\[2pt]
\textbf{Round 1, agent 1:} FINAL: 6 \dots\ ``Carson needs to heat the house from 12 to 32 degrees, a total of 20 degrees. Since each log heats the house by 5 degrees, Carson needs 20 / 5 = 4 logs to heat the house to 32 degrees. However, this calculation \dots''\\
\textbf{Round 1, agent 2:} FINAL: 6 \dots\ ``To reach 32 degrees, Carson needs to heat the house by 12 degrees. \dots\ Carson needs 12 / 5 = 2.4 logs \dots''\\
\textbf{Round 2, both agents:} FINAL: 4, with the rationale of agent 1 (``from 12 to 32 degrees, a total of 20 degrees \dots\ 20 / 5 = 4 logs''); unchanged in Round~3.
\end{quote}
  The two agents fail for different reasons.  Agent~1 commits to an incorrect answer before reasoning and then computes the correct value in its own rationale; agent~2 makes a reasoning error and, in Round~2, adopts the reasoning of agent~1.  The example therefore shows both mechanisms that can underlie the deliberation gain in Section~\ref{sec:scaling_dynamics}: revising an answer in light of one's own reasoning, and adopting a peer's correct argument.

\paragraph{Compensatory example (qwen2.5-7b, RealFP, $N=5$, first run).}  Selection rule: label-clean item, all five Round-1 estimates within one order of magnitude of each other, and a geometric mean more than one order of magnitude from the reference (115 items).
 \begin{quote}\small
\textbf{Question:} If all the people in Australia joined hands and stretched themselves out in a straight line, how long would it reach? (Give your answer as a single number in meters.) (Reference: $4.57\times10^{7}$.)\\[2pt]
\textbf{Round 1, agent 1:} FINAL: 2.5e6 \dots\ ``$2.5 \times 10^7$ people $\times$ 1.5 meters/person $= 3.75 \times 10^7$ meters \dots\ $3.75 \times 10^7$ meters $\times\, 0.8 = 3 \times 10^6$ meters''\\
\textbf{Round 1, agents 2, 3, and 5:} FINAL: 2.5e6, 3e6, and 4e6.\\
\textbf{Round 1, agent 4:} FINAL: 2.5e7 \dots\ ``$2.6 \times 10^7$ people $\times$ 1.7 meters/person $= 4.42 \times 10^7$ meters''\\
\textbf{Round 3, agent 4:} FINAL: 4e6 \dots\ ``since the majority of the peer estimates are around $4 \times 10^6$ meters \dots\ my revised estimate is $2.5 \times 10^7$ people $\times$ 1.5 meters/person $/ 10 = 3.75 \times 10^6$ meters''
\end{quote}
 All agents use the same decomposition (population $\times$ arm span), and four of the five report an answer about ten times smaller than the product computed in their own rationale, so the geometric mean is one order of magnitude below the reference and inherits a shared item-level bias $b_k$ (Section~\ref{sec:compensatory_ceiling}).  During revision, the agent whose estimate was closest to the reference divides its own correct result by ten to match the majority, although the revision prompt explicitly states that consensus is not the goal; deliberation thus increases the error of the team.

\FloatBarrier

\section{Additional Scaling and Disjunctive Results}
\label{app:disj}

\begin{figure}[h]
\centering
\includegraphics[width=0.42\textwidth]{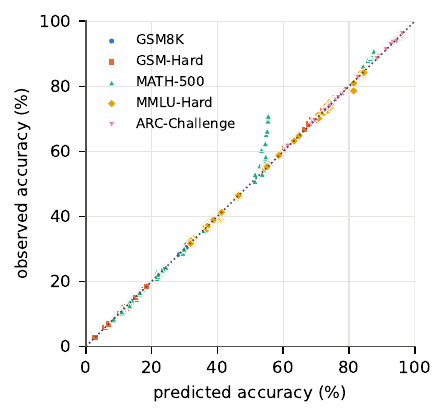}
\caption{Observed Round-1 plurality accuracy against the accuracy predicted by resampling from the estimated item-level answer distributions, for 13 models, five tasks, and all team sizes (650 points). The dotted line is the identity. Marker shape and colour denote the task; each point is one model--task--team-size configuration.}
\label{fig:pred_obs}
\end{figure}

\begin{figure}[h]
\centering
\includegraphics[width=0.42\textwidth]{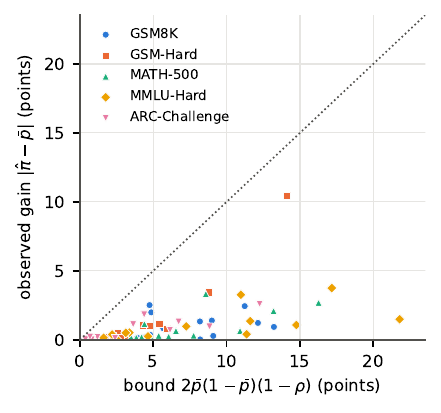}
\caption{Validation of the voting-gain bound (Eq.~\ref{eq:voting_bound}). Observed gain of the predicted large-team plurality limit over mean single-agent accuracy, $|\hat\pi-\bar p|$, against the bound $2\bar p(1-\bar p)(1-\rho)$, for the 13 panel models and five disjunctive tasks (65 points). The dotted line is the identity; every point lies below it.}
\label{fig:eq3}
\end{figure}

 This appendix reports the exact values underlying Figure~\ref{fig:scaling} (Table~\ref{tab:task_type_scaling}), the validation of the conditional-independence model of Proposition~\ref{thm:scaling_divergence} (Figures~\ref{fig:pred_obs} and~\ref{fig:eq3}), a temperature sweep (Table~\ref{tab:temp_sweep_disj}), and the statistics of the deliberation peak (Table~\ref{tab:disjunctive_ceiling_stats}).

\begin{table}[h]
\centering
\caption{\textbf{Empirical Scaling Across Disjunctive and Compensatory Tasks.} Mean accuracy (\%) across the 13-model panel, comparing Round~1 (independent answers, plurality vote or geometric mean), the Round-1 oracle success rate pass@$N$ (at least one agent correct), and Round~3 (after two revision rounds) across team sizes $N=2$ to $N=30$. The last column gives the largest half-width of the 95\% item-bootstrap interval across team sizes in each row.}
\label{tab:task_type_scaling}
\begin{center}
\small
\resizebox{\textwidth}{!}{
\begin{tabular}{llcccccccccc}
\toprule
\textbf{Task} & \textbf{Bench.} & \multicolumn{9}{c}{\textbf{Team Size } $\mathbf{N}$} & \textbf{95\% CI} \\
\cmidrule(lr){3-11}
 & & \textbf{2} & \textbf{3} & \textbf{5} & \textbf{7} & \textbf{10} & \textbf{15} & \textbf{20} & \textbf{25} & \textbf{30} & \textbf{half-width} \\
\midrule
\multicolumn{12}{l}{\textbf{Panel A: Round 1 (Independent Ensembles / Statistical Aggregation)}} \\
Disj. & arc & 82.88 & 83.42 & 83.49 & 83.66 & 83.62 & 83.70 & 83.55 & 83.65 & 83.66 & {\scriptsize$\pm$1.3} \\
Disj. & gsm8k & 34.06 & 34.16 & 34.14 & 34.34 & 34.33 & 34.25 & 34.36 & 34.27 & 34.31 & {\scriptsize$\pm$1.2} \\
Disj. & gsmhard & 18.03 & 18.35 & 18.67 & 18.73 & 18.83 & 18.96 & 18.95 & 19.05 & 19.07 & {\scriptsize$\pm$1.1} \\
Disj. & math500 & 30.80 & 31.13 & 31.31 & 31.59 & 31.74 & 31.93 & 32.07 & 32.05 & 32.08 & {\scriptsize$\pm$2.1} \\
Disj. & mmlu\_hard & 52.18 & 52.53 & 52.90 & 52.61 & 52.90 & 52.89 & 52.99 & 52.87 & 52.94 & {\scriptsize$\pm$2.0} \\
\rowcolor{gray!15} Comp. & fermi* & 31.15 & 31.81 & 32.89 & 33.12 & 32.82 & 32.91 & 33.33 & 33.14 & 33.35 & {\scriptsize$\pm$2.5} \\
\rowcolor{gray!15} Comp. & fermi+ & 1.77 & 1.76 & 1.73 & 1.73 & 1.72 & 1.72 & 1.71 & 1.72 & 1.71 & {\scriptsize$\pm$0.07} \\
\rowcolor{gray!15} Comp. & fermi-clean* & 40.62 & 41.48 & 43.00 & 43.35 & 42.94 & 42.96 & 43.53 & 43.27 & 43.61 & {\scriptsize$\pm$2.7} \\
\rowcolor{gray!15} Comp. & fermi-clean+ & 1.44 & 
1.42 & 
1.39 & 
1.39 & 
1.38 & 
1.38 & 
1.37 & 
1.37 & 
1.37 & {\scriptsize$\pm$0.06} \\
\midrule
\multicolumn{12}{l}{\textbf{Panel B: Round 1 Oracle Success (pass@$N$)}} \\
Disj. & arc & 84.82 & 85.83 & 86.61 & 87.12 & 87.46 & 87.90 & 88.14 & 88.26 & 88.43 & {\scriptsize$\pm$1.2} \\
Disj. & gsm8k & 37.79 & 39.93 & 43.31 & 45.47 & 47.83 & 50.32 & 52.17 & 53.47 & 54.44 & {\scriptsize$\pm$1.2} \\
Disj. & gsmhard & 20.32 & 21.62 & 23.22 & 24.29 & 25.47 & 26.92 & 27.95 & 28.41 & 28.87 & {\scriptsize$\pm$1.5} \\
Disj. & math500 & 33.64 & 35.80 & 37.97 & 39.87 & 41.65 & 43.67 & 44.91 & 46.25 & 47.08 & {\scriptsize$\pm$2.1} \\
Disj. & mmlu\_hard & 56.50 & 58.86 & 61.51 & 63.21 & 64.58 & 66.05 & 66.89 & 67.28 & 67.67 & {\scriptsize$\pm$1.9} \\
\rowcolor{gray!15} Comp. & fermi* & 37.70 & 41.24 & 46.37 & 48.41 & 51.09 & 53.48 & 55.10 & 56.37 & 57.32 & {\scriptsize$\pm$2.7} \\
\rowcolor{gray!15} Comp. & fermi-clean* & 48.39 & 52.80 & 58.72 & 60.96 & 63.86 & 66.67 & 68.47 & 69.48 & 70.80 & {\scriptsize$\pm$2.7} \\
\midrule
\multicolumn{12}{l}{\textbf{Panel C: Round 3 (Deliberated Teams / Multi-Turn Communication)}} \\
Disj. & arc & 84.05 & 84.45 & 84.42 & 84.40 & 84.43 & 84.53 & 84.54 & 84.55 & 84.54 & {\scriptsize$\pm$1.3} \\
Disj. & gsm8k & 60.51 & 61.11 & 61.90 & 61.95 & 61.76 & 61.70 & 61.62 & 61.35 & 60.88 & {\scriptsize$\pm$1.2} \\
Disj. & gsmhard & 32.21 & 32.47 & 32.89 & 33.15 & 33.36 & 33.14 & 33.53 & 33.14 & 33.08 & {\scriptsize$\pm$1.8} \\
Disj. & math500 & 38.96 & 38.87 & 39.66 & 39.75 & 39.43 & 39.90 & 39.40 & 39.36 & 39.13 & {\scriptsize$\pm$2.3} \\
Disj. & mmlu\_hard & 54.52 & 54.70 & 55.18 & 55.05 & 54.95 & 54.98 & 54.89 & 54.89 & 54.81 & {\scriptsize$\pm$2.0} \\
\rowcolor{gray!15} Comp. & fermi* & 30.57 & 31.76 & 32.55 & 32.72 & 32.70 & 33.04 & 33.21 & 33.16 & 34.01 & {\scriptsize$\pm$2.5} \\
\rowcolor{gray!15} Comp. & fermi+ & 1.78 & 1.76 & 1.74 & 1.73 & 1.73 & 1.74 & 1.73 & 1.73 & 1.72 & {\scriptsize$\pm$0.07} \\
\rowcolor{gray!15} Comp. & fermi-clean* & 40.18 & 41.51 & 42.48 & 42.81 & 42.92 & 43.34 & 43.51 & 43.25 & 44.61 & {\scriptsize$\pm$2.8} \\
\rowcolor{gray!15} Comp. & fermi-clean+ & 1.45 & 
1.43 & 
1.40 & 
1.39 & 
1.39 & 1.39 & 1.39 & 1.39 & 1.37 & {\scriptsize$\pm$0.06} \\
\bottomrule
\end{tabular}
}
\end{center}
\begin{flushleft}
\scriptsize
* Accuracy within one order of magnitude (\%); + mean absolute $\log_{10}$ error (clipped at $\pm3$).  Fermi results use the decomposition-prompt runs of ten models (Appendix~\ref{sec:integrity}); fermi-clean is restricted to the 385 items whose reference value lies within three orders of magnitude of the median model estimate (Section~\ref{sec:label_noise}).
\end{flushleft}
\end{table}
\FloatBarrier

\FloatBarrier

\paragraph{Prediction of plurality accuracy.}  Figure~\ref{fig:pred_obs} tests the disjunctive part of Proposition~\ref{thm:scaling_divergence}.  For each model and task, we estimate the item-level answer distribution $\hat P_k$ from all Round-1 answers of the $N=30$ teams (up to 90 samples per item across runs).  For every team size $N$, we then draw $N$ answers independently from $\hat P_k$ (200 draws per item), apply plurality voting with random tie-breaking, and average the resulting accuracy over items.  The prediction thus assumes only that agents are conditionally independent given the item; it uses no information from the teams of size $N<30$, whose answers were generated in separate runs and therefore constitute an out-of-sample test; only the $N=30$ points are in-sample.  Across 650 configurations (13 models, five tasks, and ten team sizes), the mean absolute difference between predicted and observed accuracy is 0.48 points (median 0.18; 95th percentile 1.17; Pearson $r=0.999$).  The largest deviations occur on MATH-500 for gpt-oss-20b, r1-distill-qwen-14b, and qwen3-8b, where observed accuracy exceeds the prediction: the evaluation pipeline merges mathematically equivalent answers written in different forms before voting, whereas $\hat P_k$ is estimated over answer strings.  Excluding MATH-500, the mean absolute difference is 0.23 points and the maximum is 2.8 points.

\FloatBarrier

\paragraph{Validation of the voting-gain bound.} Figure~\ref{fig:eq3} compares, for every panel model and disjunctive task, the observed gain $|\hat\pi-\bar p|$ with the bound of Eq.~\ref{eq:voting_bound}, computed from the mean single-agent success $\bar p$ and the intra-class correlation $\rho$ of the same model and task.  The bound holds in all 65 cases.  It is also loose: averaged per task, the observed gain amounts to 11--21\% of the bound, because the bound assumes a binary outcome, whereas incorrect answers on these tasks are often spread over several values.  The bound is proven for a single dominant incorrect answer; for tasks with many answer options, Figure~\ref{fig:eq3} therefore provides an empirical rather than a formal confirmation.

\paragraph{Sampling temperature on disjunctive tasks.}  Table~\ref{tab:temp_sweep_disj} reports a temperature sweep for qwen2.5-7b with the answer-first prompts of the main experiments on GSM-Hard (400 items), MMLU-Hard (400 items), and GPQA Diamond \citep{rein2024gpqa} (198 items), with one run per temperature.  Raising the temperature from 0.2 to 1.0 increases the dispersion of answers within a team and lowers the intra-class correlation of correctness on every task, and pass@30 rises accordingly (by 9.3, 8.3, and 14.1 points).  Plurality accuracy at $N=30$, however, remains within 2.3 points of single-agent accuracy at every temperature, and the predicted large-team limit (not shown) agrees with it to within 1 point.  The additional dispersion therefore falls on incorrect answers and leaves the modal answer of most items unchanged: a lower $\rho$ raises the bound of Eq.~\ref{eq:voting_bound} but does not by itself realise it, and the process loss of plurality voting (pass@30 minus plurality accuracy) grows with temperature.  Round-3 accuracy is also largely insensitive to temperature.

\begin{table}[h] 
\caption{Temperature sweep on disjunctive tasks (qwen2.5-7b, answer-first prompts, one run). Accuracy in \%. ``Gain'' is Round-1 plurality accuracy at $N=30$ minus single-agent accuracy, with a paired 95\% item-bootstrap interval; $\rho$ is the intra-class correlation of correctness; ``Distinct'' is the mean number of distinct answers among the 30 Round-1 answers to an item.}
\label{tab:temp_sweep_disj}
\begin{center}
\footnotesize
\setlength{\tabcolsep}{3.5pt}
\resizebox{0.9\textwidth}{!}{
\begin{tabular}{lcccccccc}
\toprule
\textbf{Task} & $T$ & \textbf{Solo} & \textbf{R1, $N{=}30$} & \textbf{Gain} & \textbf{pass@30} & $\bm{\rho}$ & \textbf{Distinct} & \textbf{R3, $N{=}30$} \\
\midrule
GSM-Hard & 0.2 & 11.0 & 10.0 & $-1.0$\,{\scriptsize[-2.0, -0.2]} & 13.5 & 0.92 & 3.3 & 27.0 \\
 & 0.4 & 11.0 & 10.8 & $-0.2$\,{\scriptsize[-1.5, +1.0]} & 18.2 & 0.85 & 6.2 & 28.0 \\
 & 0.7 & 11.0 & 10.8 & $-0.2$\,{\scriptsize[-2.0, +1.5]} & 19.8 & 0.79 & 10.0 & 27.3 \\
 & 1.0 & 10.2 & 11.0 & $+0.8$\,{\scriptsize[-1.0, +2.8]} & 22.8 & 0.75 & 13.0 & 29.0 \\
\addlinespace
MMLU-Hard & 0.2 & 54.5 & 53.2 & $-1.2$\,{\scriptsize[-3.0, +0.2]} & 55.5 & 0.98 & 1.0 & 58.0 \\
 & 0.4 & 55.2 & 53.0 & $-2.2$\,{\scriptsize[-4.2, -0.5]} & 58.2 & 0.95 & 1.1 & 57.0 \\
 & 0.7 & 54.8 & 53.2 & $-1.5$\,{\scriptsize[-3.8, +0.5]} & 61.3 & 0.90 & 1.2 & 56.5 \\
 & 1.0 & 54.8 & 53.0 & $-1.8$\,{\scriptsize[-4.0, +0.8]} & 63.7 & 0.87 & 1.3 & 57.0 \\
\addlinespace
GPQA Diamond & 0.2 & 29.3 & 30.8 & $+1.5$\,{\scriptsize[+0.0, +3.5]} & 33.3 & 0.97 & 1.1 & 32.3 \\
 & 0.4 & 30.3 & 30.3 & $+0.0$\,{\scriptsize[-2.0, +2.0]} & 37.4 & 0.93 & 1.2 & 30.3 \\
 & 0.7 & 29.8 & 29.3 & $-0.5$\,{\scriptsize[-2.5, +1.5]} & 42.9 & 0.85 & 1.4 & 30.3 \\
 & 1.0 & 30.8 & 29.3 & $-1.5$\,{\scriptsize[-4.5, +1.5]} & 47.5 & 0.80 & 1.6 & 30.3 \\
\bottomrule
\end{tabular}
}
\end{center}
\end{table}
\FloatBarrier

\begin{table}[h]
\caption{\textbf{Anatomy of the Disjunctive Deliberation Peak and Decay.} Mean accuracy (\%) across the 13-model panel: Round~1 at $N=2$ and $N=30$, Round~3 optimal team size $N^*$, peak accuracy, accuracy at $N=30$, post-peak change $\Delta_{\text{decay}} = \text{Acc}_{N=30} - \text{Acc}_{N^*}$, and net deliberation gain at the peak $\Delta_{\text{delib}} = \text{Acc}_{N^*} - \text{Acc}_{\text{R1}, N=2}$, each with a paired 95\% item-bootstrap interval. Because $N^*$ is selected as the maximum, $\Delta_{\text{decay}}$ is biased downward.}
\label{tab:disjunctive_ceiling_stats}
\begin{center}
\footnotesize
\resizebox{\textwidth}{!}{
\begin{tabular}{lccccccc}
\toprule
\textbf{Bench-} & \textbf{R1} & \textbf{R1} & $\mathbf{N^*}$ & \textbf{R3} & \textbf{R3} & $\mathbf{\Delta_{\text{decay}}}$ & $\mathbf{\Delta_{\text{delib}}}$\\
\textbf{mark} & \textbf{$N=2$} & \textbf{$N=30$} & \textbf{Peak} & & \textbf{$N=30$} &  &  \\
\midrule
arc & 82.88 & 83.66 & 25 & \textbf{84.55} & 84.54 & $-0.01$\,{\scriptsize[-0.13, +0.13]} & $+1.66$\,{\scriptsize[+1.38, +1.95]} \\
gsm8k & 34.06 & 34.31 & 7 & \textbf{61.95} & 60.88 & $-1.07$\,{\scriptsize[-1.39, -0.73]} & $+27.89$\,{\scriptsize[+27.01, +28.79]} \\
gsmhard & 18.03 & 19.07 & 20 & \textbf{33.53} & 33.08 & $-0.45$\,{\scriptsize[-0.69, -0.21]} & $+15.50$\,{\scriptsize[+14.49, +16.52]} \\
math500 & 30.80 & 32.08 & 15 & \textbf{39.90} & 39.13 & $-0.77$\,{\scriptsize[-1.22, -0.33]} & $+9.10$\,{\scriptsize[+8.08, +10.08]} \\
mmlu\_hard & 52.18 & 52.94 & 5 & \textbf{55.18} & 54.81 & $-0.37$\,{\scriptsize[-0.73, -0.01]} & $+3.00$\,{\scriptsize[+2.42, +3.58]} \\
\bottomrule
\end{tabular}
}
\end{center}
\end{table}
\FloatBarrier

\section{Additional Compensatory Results}
\label{app:comp}

\begin{figure}[h] 
\centering
\includegraphics[width=\textwidth]{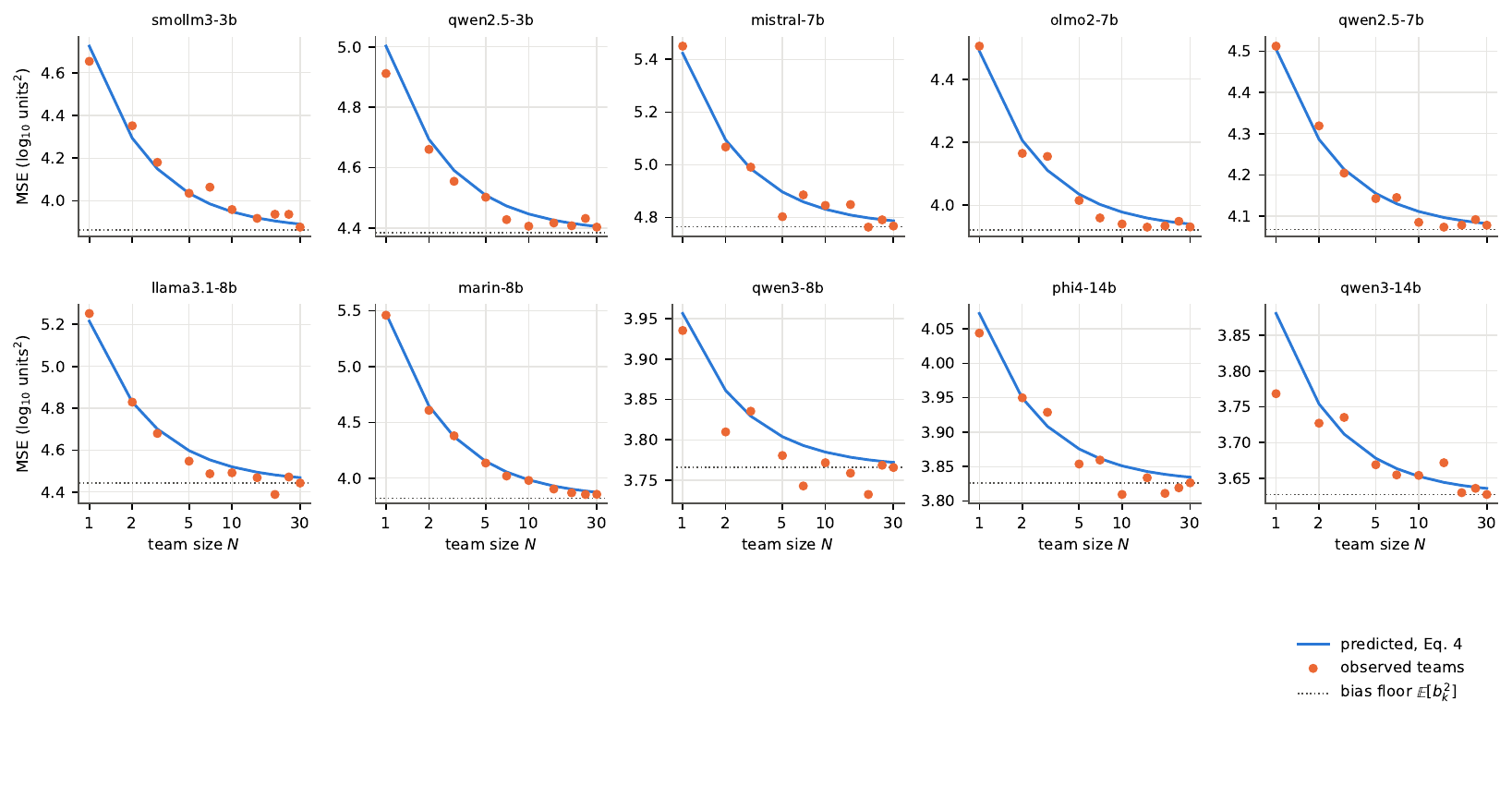}
\caption{Validation of the compensatory decomposition (Eq.~\ref{eq:comp_mse}). For each panel model, the mean squared error of the geometric mean of clipped log-errors predicted from item-level bias and within-item variance (line) and observed for single agents and actual Round-1 teams (points), together with the bias floor $\mathbb{E}_k[b_k^2]$ (dotted). Decomposition-prompt runs. The $y$-axes do not start at zero, so the vertical scale exaggerates the decrease; the observed reduction in mean squared error from $N=1$ to $N=30$ is at most 13\% for seven of the ten models and 15\%, 17\%, and 29\% for llama3.1-8b, smollm3-3b, and marin-8b, respectively.}
\label{fig:eq4}
\end{figure}

 This appendix reports the effect of sampling temperature (Table~\ref{tab:temperature_grid}), the distribution of signed errors of single agents (Table~\ref{tab:unbiased_stats}), and the cross-model correlation of item-level biases (Table~\ref{tab:cross_model_corr}), which are discussed in Section~\ref{sec:compensatory_ceiling}, and it begins with a direct validation of Eq.~\ref{eq:comp_mse} (Figure~\ref{fig:eq4}) and a sensitivity analysis of the clipping threshold (Table~\ref{tab:clip_sensitivity}).

\paragraph{Validation of the compensatory decomposition.}  For each model, we estimate the item-level bias $b_k$ and the within-item variance $\sigma_k^2$ of the clipped log-errors from the Round-1 answers of the $N=30$ teams, and predict the mean squared error of the geometric mean as $\mathbb{E}_k[b_k^2]+\mathbb{E}_k[\sigma_k^2]/N$.  We compare this prediction with the observed mean squared error of single agents and of the actual Round-1 teams of each size, which were generated in separate runs; only the $N=30$ points are in-sample.  For all ten models, the mean relative difference between predicted and observed error is 0.6\% (maximum 3.0\%; Pearson $r=0.997$ over 90 out-of-sample points).  The curves in Figure~\ref{fig:eq4} flatten towards the bias floor $\mathbb{E}_k[b_k^2]$ within five to ten agents, which is the compensatory ceiling predicted by Proposition~\ref{thm:scaling_divergence}.

\FloatBarrier

\paragraph{Sensitivity to the clipping threshold.}  Log-errors are clipped at $\pm3$ throughout, which limits the influence of implausible reference values (Section~\ref{sec:label_noise}) and of occasional estimates that are wrong by tens of orders of magnitude.  Table~\ref{tab:clip_sensitivity} repeats the main compensatory analyses with thresholds of $\pm2$ and $\pm5$ and without clipping.  At every threshold, averaging reduces the error by only 4--8\%, and item-level bias accounts for 86--94\% of the mean squared error.  Without clipping, the mean is dominated by a small number of extreme items, which raises both the bias share and the cross-model correlation; clipping therefore makes the reported ceiling conservative rather than producing it.  The median absolute log-error, which requires no clipping, decreases only from $2.04$ for a single agent to $1.96$ at $N=30$.

\begin{table}[h]
\caption{\textbf{Sensitivity of the RealFP Results to Clipping.} Mean absolute log-error (MAE) of the geometric mean in Round~1 (single agents at $N=1$), its relative reduction from $N=1$ to $N=5$ and to $N=30$, bias share $\beta=\mathbb{E}[b_k^2]/(\mathbb{E}[b_k^2]+\mathbb{E}[\sigma_k^2])$, and mean pairwise cross-model correlation of item-level biases $r$, macro-averaged over the ten models with decomposition-prompt runs.  The $\pm3$ row corresponds to the values reported in the main text.}
\label{tab:clip_sensitivity}
\begin{center}
\footnotesize
\begin{tabular}{lccccccc}
\toprule
\textbf{Clipping} & \textbf{MAE} & \textbf{MAE} & \textbf{MAE} & \textbf{Red.} & \textbf{Red.} & $\beta$ & $r$ \\
 & $\mathbf{N=1}$ & $\mathbf{N=5}$ & $\mathbf{N=30}$ & \textbf{1$\to$5} & \textbf{1$\to$30} & & \\
\midrule
$\pm2$ & 1.41 & 1.32 & 1.30 & 6.9\% & 7.9\% & 0.86 & 0.61 \\
$\pm3$ & 1.84 & 1.73 & 1.71 & 6.1\% & 7.1\% & 0.87 & 0.63 \\
$\pm5$ & 2.41 & 2.28 & 2.26 & 5.5\% & 6.4\% & 0.88 & 0.65 \\
none   & 3.82 & 3.68 & 3.66 & 3.7\% & 4.2\% & 0.94 & 0.84 \\
\bottomrule
\end{tabular}
\end{center}
\end{table}
\FloatBarrier

\paragraph{Sampling temperature.}  Table~\ref{tab:temperature_grid} reports Round-1 within-one-order-of-magnitude accuracy (median aggregation) on all 529 RealFP items at three temperatures, after correcting the parsing of exponents written with superscript digits (Appendix~\ref{sec:integrity}).  Accuracy is essentially invariant to temperature: all differences between temperatures lie well within the 95\% intervals of about $\pm4$ points.  Higher temperature does increase the dispersion of estimates within a team: the within-item standard deviation of log-errors grows by 30--70\% from $T=0.4$ to $T=1.0$, and the share of unanimous teams falls from 33\% to 4\%.  The between-item standard deviation remains several times larger, and the intra-class correlation remains between 0.83 and 0.98, so additional sampling diversity does not translate into better aggregates.

\begin{table}[h]
\caption{Within-one-order-of-magnitude accuracy (\%) on RealFP by temperature and team size (Round~1, median aggregation, 529 items, one to three runs per cell); $\pm$ gives the half-width of the 95\% item-bootstrap interval.}
\label{tab:temperature_grid}
\begin{center}
\small
\setlength{\tabcolsep}{4pt}
\begin{tabular}{lccccccccc}
\toprule
& \multicolumn{3}{c}{$T=0.4$} & \multicolumn{3}{c}{$T=0.8$} & \multicolumn{3}{c}{$T=1.0$} \\
\cmidrule(lr){2-4}\cmidrule(lr){5-7}\cmidrule(lr){8-10}
\textbf{Model} & $N{=}2$ & $N{=}5$ & $N{=}10$ & $N{=}2$ & $N{=}5$ & $N{=}10$ & $N{=}2$ & $N{=}5$ & $N{=}10$ \\
\midrule
qwen3-4b & 34.3{\tiny$\pm$3.9} & 35.5{\tiny$\pm$3.7} & 35.3{\tiny$\pm$3.6} & 33.5{\tiny$\pm$3.6} & 34.7{\tiny$\pm$3.9} & 35.8{\tiny$\pm$3.8} & 33.4{\tiny$\pm$3.4} & 35.2{\tiny$\pm$3.8} & 36.2{\tiny$\pm$3.9} \\
qwen3-8b & 38.3{\tiny$\pm$4.0} & 38.4{\tiny$\pm$3.9} & 38.5{\tiny$\pm$4.1} & 39.4{\tiny$\pm$3.7} & 38.6{\tiny$\pm$3.9} & 38.7{\tiny$\pm$3.7} & 39.2{\tiny$\pm$3.9} & 38.8{\tiny$\pm$3.9} & 38.9{\tiny$\pm$4.3} \\
qwen3-14b & 38.1{\tiny$\pm$4.0} & 38.2{\tiny$\pm$4.2} & 38.3{\tiny$\pm$4.4} & 38.1{\tiny$\pm$3.7} & 37.6{\tiny$\pm$4.1} & 38.9{\tiny$\pm$4.2} & 38.9{\tiny$\pm$3.9} & 38.0{\tiny$\pm$4.0} & 38.4{\tiny$\pm$4.1} \\
gpt-oss-20b & 50.3{\tiny$\pm$3.9} & 55.3{\tiny$\pm$3.9} & 55.2{\tiny$\pm$4.0} & 49.9{\tiny$\pm$3.7} & 55.0{\tiny$\pm$3.9} & 55.4{\tiny$\pm$4.0} & 51.0{\tiny$\pm$3.8} & 56.1{\tiny$\pm$3.9} & 54.8{\tiny$\pm$4.2} \\
\bottomrule
\end{tabular}
\end{center}
\end{table}

\begin{table}[h] 
\caption{\textbf{Intra-Model Bias and Distribution of Signed Errors.} Mean signed error $b = \mathbb{E}[e_i]$ (clipped at $\pm3$; 95\% item-cluster bootstrap interval), median signed error, standard deviation $\sigma$ (clipped), and percentage of estimates under- and over-estimating the ground truth. Evaluated on single-agent estimates of the decomposition-prompt runs, sorted by parameter count.}
\label{tab:unbiased_stats}
\begin{center}
\small
\begin{tabular}{lcccccc}
\toprule
\textbf{Model} & \textbf{Obs.} & $\mathbf{\mathbb{E}[e_i]}$ \textbf{(Mean }$\mathbf{b}$\textbf{)} & \textbf{Median} & $\bm{\sigma}$ & \textbf{Under} & \textbf{Over} \\
 & & \textbf{(OOM)} & \textbf{(OOM)} & \textbf{(Std)} & \textbf{(\%)} & \textbf{(\%)} \\
\midrule
smollm3-3b & 1,058 & $+0.48$\,{\scriptsize[+0.32, +0.66]} & $+0.49$ & 2.10 & 41.2 & 57.9 \\
qwen2.5-3b & 1,058 & $+0.65$\,{\scriptsize[+0.47, +0.84]} & $+0.92$ & 2.12 & 36.1 & 63.5 \\
mistral-7b & 522 & $+0.71$\,{\scriptsize[+0.53, +0.89]} & $+1.21$ & 2.23 & 35.6 & 63.6 \\
olmo2-7b & 1,158 & $+0.19$\,{\scriptsize[+0.02, +0.37]} & $+0.28$ & 2.11 & 45.1 & 54.5 \\
qwen2.5-7b & 1,587 & $+0.11$\,{\scriptsize[-0.05, +0.28]} & $+0.02$ & 2.12 & 48.9 & 50.1 \\
llama3.1-8b & 528 & $+0.62$\,{\scriptsize[+0.42, +0.79]} & $+0.97$ & 2.21 & 39.0 & 60.8 \\
marin-8b & 1,585 & $-0.20$\,{\scriptsize[-0.38, -0.03]} & $-0.19$ & 2.33 & 52.2 & 47.1 \\
qwen3-8b & 525 & $-0.04$\,{\scriptsize[-0.20, +0.13]} & $-0.03$ & 1.99 & 50.5 & 48.2 \\
phi4-14b & 528 & $+0.47$\,{\scriptsize[+0.31, +0.64]} & $+0.35$ & 1.96 & 39.8 & 58.7 \\
qwen3-14b & 529 & $+0.16$\,{\scriptsize[+0.00, +0.31]} & $+0.08$ & 1.94 & 46.1 & 52.0 \\
\bottomrule
\end{tabular}
\end{center}
\end{table}

\begin{table}[h] 
\caption{\textbf{Cross-Model Correlation of Item-Level Biases ($\mathbf{r}$).} Pairwise Pearson correlation of item-level mean signed log-errors (clipped at $\pm3$) across the ten models with decomposition-prompt runs, on the items with estimates from all models, sorted by parameter count. The mean off-diagonal correlation is 0.63 (95\% item-bootstrap interval [0.60, 0.67]).}
\label{tab:cross_model_corr}
\begin{center}
\footnotesize
\setlength{\tabcolsep}{1.6pt}
\begin{tabular}{lcccccccccc}
\toprule
 & \rotatebox{90}{smollm3-3b} & \rotatebox{90}{qwen2.5-3b} & \rotatebox{90}{mistral-7b} & \rotatebox{90}{olmo2-7b} & \rotatebox{90}{qwen2.5-7b} & \rotatebox{90}{llama3.1-8b} & \rotatebox{90}{marin-8b} & \rotatebox{90}{qwen3-8b} & \rotatebox{90}{phi4-14b} & \rotatebox{90}{qwen3-14b} \\
\midrule
smollm3-3b & 1.00 & 0.69 & 0.55 & 0.66 & 0.64 & 0.64 & 0.65 & 0.62 & 0.58 & 0.64 \\
qwen2.5-3b & 0.69 & 1.00 & 0.50 & 0.68 & 0.67 & 0.64 & 0.60 & 0.64 & 0.59 & 0.64 \\
mistral-7b & 0.55 & 0.50 & 1.00 & 0.52 & 0.53 & 0.65 & 0.55 & 0.46 & 0.59 & 0.52 \\
olmo2-7b & 0.66 & 0.68 & 0.52 & 1.00 & 0.66 & 0.61 & 0.66 & 0.66 & 0.61 & 0.66 \\
qwen2.5-7b & 0.64 & 0.67 & 0.53 & 0.66 & 1.00 & 0.69 & 0.61 & 0.66 & 0.71 & 0.67 \\
llama3.1-8b & 0.64 & 0.64 & 0.65 & 0.61 & 0.69 & 1.00 & 0.61 & 0.60 & 0.68 & 0.67 \\
marin-8b & 0.65 & 0.60 & 0.55 & 0.66 & 0.61 & 0.61 & 1.00 & 0.63 & 0.61 & 0.65 \\
qwen3-8b & 0.62 & 0.64 & 0.46 & 0.66 & 0.66 & 0.60 & 0.63 & 1.00 & 0.73 & 0.77 \\
phi4-14b & 0.58 & 0.59 & 0.59 & 0.61 & 0.71 & 0.68 & 0.61 & 0.73 & 1.00 & 0.72 \\
qwen3-14b & 0.64 & 0.64 & 0.52 & 0.66 & 0.67 & 0.67 & 0.65 & 0.77 & 0.72 & 1.00 \\
\bottomrule
\end{tabular}
\end{center}
\end{table}
\FloatBarrier

\section{De-Saturating Compensatory LLMs: Interventions and Specialist Models}
\label{sec:interventions}

  Having established that homogeneous LLM teams do not overcome the compensatory ceiling on their own, we explore structured interventions intended to decorrelate errors ($\rho \to 0$) or to reduce systematic bias ($b^2 \to 0$).

\subsection{The Heterogeneous Intervention Battery}
  {\sloppy We evaluate interventions on a heterogeneous team of five 7B--8B models (qwen2.5-7b, llama3.1-8b, mistral-7b, olmo2-7b, and marin-8b; one run, $T=0.4$, 529 items):\par}
\begin{enumerate}
    \item   \textbf{Heterogeneous Baseline:} the five models without intervention.
    \item   \textbf{Prefix Anchoring:} each agent receives a different decomposition strategy (e.g., top-down, bottom-up, rate $\times$ time) as a prompt prefix.
    \item   \textbf{Structured Scaffolding:} each agent follows a different complete estimation template.
    \item   \textbf{Log-Calculator Offloading (logcalc):} agents list estimated factors, and the answer is computed from them in log space.
    \item   \textbf{Unit System Change-of-Basis:} agents reason in different unit systems, combined with logcalc.
    \item   \textbf{Constant Grounding:} a table of reference constants is inserted into the prompt, combined with logcalc.
\end{enumerate}

From the results  reported in Table~\ref{tab:interventions} it can be seen that Structured scaffolding yields the highest accuracy at $N=5$ ($36.3\%$, 95\% CI $[32.1, 40.6]$, compared with $33.1\%$, $[29.3, 36.9]$, for the baseline) and the lowest mean log-error ($3.31$, unclipped), although its interval overlaps with that of the baseline. Every logcalc variant reduces accuracy relative to the baseline. The logcalc and grounding conditions at $N=5$ were completed for only 180 and 57 items, respectively, and are not comparable with the other cells. Among homogeneous runs, logcalc also reduced accuracy for gpt-oss-20b (from $52.0\%$ to $48.0\%$ for a single agent), and decomposition anchors had no discernible effect.

\begin{table}[t]
\caption{Heterogeneous-team interventions on RealFP (Round~1, median aggregation). Accuracy within one order of magnitude (\%) and mean absolute $\log_{10}$ error (unclipped, and therefore not comparable with the clipped values of Section~\ref{sec:compensatory_ceiling}; see Table~\ref{tab:clip_sensitivity}) at $N=5$, with 95\% item-bootstrap intervals at $N=5$. Where fewer than 529 items were completed, $n$ is given.}
\label{tab:interventions}
\begin{center}
\small
\begin{tabular}{lcccc}
\toprule
\textbf{Intervention} & Acc.\ $N{=}2$ & Acc.\ $N{=}3$ & Acc.\ $N{=}5$ & Error $N{=}5$ \\
\midrule
Baseline (no intervention) & 28.9 & 31.4 & 33.1\,{\scriptsize[29.3, 36.9]} & 3.41\,{\scriptsize[2.91, 3.96]} \\
Decomposition anchors & 28.0 & 30.4 & 33.1\,{\scriptsize[29.3, 37.1]} & 3.44\,{\scriptsize[2.98, 3.96]} \\
Unit systems + logcalc & 18.0 & 20.2 & 25.0\,{\scriptsize[21.4, 28.9]} & 4.52\,{\scriptsize[3.95, 5.15]} \\
Logcalc & 18.5 & 24.4 & 28.9\,{\scriptsize[22.2, 35.6]} ($n=180$) & 5.03\,{\scriptsize[3.81, 6.35]} \\
Grounding + logcalc & 21.0 & 19.7 & 21.1\,{\scriptsize[10.5, 31.6]} ($n=57$) & 4.35\,{\scriptsize[2.96, 6.42]} \\
Distinct scaffolds & 30.2 & 33.6 & 36.3\,{\scriptsize[32.1, 40.6]} & 3.31\,{\scriptsize[2.83, 3.88]} \\
Anchors + logcalc & 18.9 & 19.7 & 26.1\,{\scriptsize[22.5, 29.9]} & 4.51\,{\scriptsize[3.97, 5.13]} \\
\bottomrule
\end{tabular}
\end{center}
\end{table}

\subsection{Evaluation of a Specialist Model}
  We also compared Ornith-1.5-9B, a publicly available model fine-tuned from Qwen3.5-9B with reinforcement learning for agentic coding, with its base model Qwen3.5-9B on RealFP and on a Science Olympiad estimation set (SciOly, 500 items), using Program-of-Thought (PoT) and decomposition-scaffold prompting.  The control runs with direct and chain-of-thought prompting used a smaller token budget and were truncated on about half of the items, so we report only the paired comparison on items that both models answered with the same 16k-token budget.  The fine-tuned model is not more accurate on RealFP ($+0.4$ points for PoT, 95\% CI $[-3.6, 4.4]$; $+0.4$ for the scaffold, $[-3.9, 4.9]$) or with PoT on SciOly ($+0.7$, $[-2.6, 3.7]$), and is more accurate only with the scaffold on SciOly ($+5.7$, $[0.7, 10.4]$).  Program-of-Thought prompting yields the highest accuracy for both models (about $50\%$ within one order of magnitude on RealFP and $76\%$ on SciOly).

\end{document}